\documentclass[sensors,article,accept,moreauthors,pdftex]{Definitions/mdpi}
\usepackage[utf8]{inputenc}
\usepackage[T1]{fontenc}
\usepackage{textcomp}
\usepackage{amsmath,amsfonts,amssymb}
\usepackage{graphicx}
\usepackage{setspace}
\usepackage{tocloft}
\usepackage{subcaption}
\usepackage{float}
\usepackage{multicol}
\usepackage{lineno}
\newcommand{\norm}[1]{\left\lVert#1\right\rVert}
\DeclareMathOperator*{\argmin}{arg\,min}
\newcommand{\R}{\rm I\!R}

\setitemize{parsep=6pt,itemsep=0pt,leftmargin=*,labelsep=5.5mm}
\setenumerate{parsep=6pt,itemsep=0pt,leftmargin=*,labelsep=5.5mm}
\setlist[description]{itemsep=0mm}   

\firstpage{1} 
\pubvolume{xx}
\issuenum{1}
\articlenumber{5}
\pubyear{2019}
\copyrightyear{2019}
\history{Received: date; Accepted: date; Published: date}

\Title{Benchmarking Deep Trackers on Aerial Videos}

\Author{Abu Md Niamul Taufique, Breton Minnehan $^{\dagger}$ and Andreas Savakis *}

\AuthorNames{Abu Md Niamul Taufique, Breton Minnehan and Andreas Savakis}

\address[1]{%
Rochester Institute of Technology, Rochester, NY 14623, USA; at7133@rit.edu (A.M.N.T.); breton.minnehan.1@us.af.mil (B.M.)}

\corres{\hangafter=1 \hangindent=1.05em \hspace{-0.82em}  {andreas.savakis@rit.edu} }

\firstnote{\hangafter=1 \hangindent=1.05em \hspace{-0.82em}  Air Force Research Laboratory, WPAFB, OH 45433, USA.} 

\abstract{
In recent years, deep learning-based visual object trackers
have achieved state-of-the-art performance on several visual object tracking benchmarks. However, most tracking benchmarks are focused on ground level videos, whereas aerial tracking presents a new set of challenges. 
In this paper, we compare ten trackers based on deep learning techniques on four aerial datasets. 
We choose top performing trackers 
utilizing different approaches, specifically tracking by detection, discriminative correlation filters, Siamese networks and reinforcement learning.
In our experiments, we use a subset of OTB2015 dataset with aerial style videos; the UAV123 dataset without synthetic sequences; the UAV20L dataset, which contains 20 long sequences; and DTB70 dataset as our benchmark datasets. 
We compare the advantages and disadvantages of different trackers in different tracking situations encountered in aerial data. 
Our findings indicate that the trackers 
perform significantly worse in aerial datasets compared to standard ground level videos. 
We attribute this effect to smaller target size, camera motion, significant camera rotation with respect to the target, out of view movement, and clutter in the form of occlusions or similar looking distractors near tracked object.
}


\begin{document}


\section{Introduction}
\label{sec:intro}
Visual object tracking is an important area of computer vision with applications in robotics~\cite{papanikolopoulos1993visual}, autonomous driving \cite{7155586, laurense2017path}, video surveillance \cite{tang2017multiple}, pose estimation \cite{girdhar2018detect}, medicine \cite{walker2017systems, speidel2014visual}, activity recognition \cite{aggarwal2014human}, and many others. Visual object tracking refers to locating a region of interest, typically a bounding box around the tracked object, in a sequence of frames.
Visual tracking is challenging due to variations in appearance, illumination and scaling, changes in zoom, rotation, distortions, occlusion, abrupt motion, similar looking distractors, out of frame movement, etc.

Early visual tracking methods relied on hand crafted features, such as optical flow, and keypoints, and related benchmarking studies analyzed their performance \cite{yilmaz2006object, smeulders2014visual, ning2009robust, zhou2009object, bolme2010visual, han2011visual, das2011particle, 
danelljan2014accurate,jia2012visual,zhong2012robust,yoon2012visual,
henriques2015high, hare2016struck, kiani2017learning, danelljan2015learning, dong2015visual, ontiveros2015objects}. 
Popular trackers such as Kernelized Correlation Filters (KCF) \cite{henriques2015high}, 
Structured output tracking with Kernels (STRUCK) \cite{hare2016struck}, 
Spatially Regularized Discriminative Correlation Filters (SRDCF) \cite{danelljan2015learning}, 
and Background-Aware Correlation Filters (BACF) \cite{kiani2017learning} 
used hand-crafted features. 
However, traditional methods may fail in challenging situations, 
such as those encountered in the 2018 Visual Object Tracking (VOT 2018) challenge \cite{kristan2018visual} and the 2015 Object Tracking Benchmark (OTB 2015) challenge~\cite{wu2015object}.
In recent years, deep learning has advanced object detection and other computer vision tasks, including tracking, semantic segmentation, pose estimation, visual question answering, and style transfer. In the VOT 2018 challenge, almost all of the top performing trackers used deep learning features based on convolutional neural networks (CNN) \cite{kristan2018visual}. In our study, we focus on trackers based on such deep learning features.

\begin{figure}[H]
\begin{center}
   \includegraphics[width= 0.9 \textwidth]{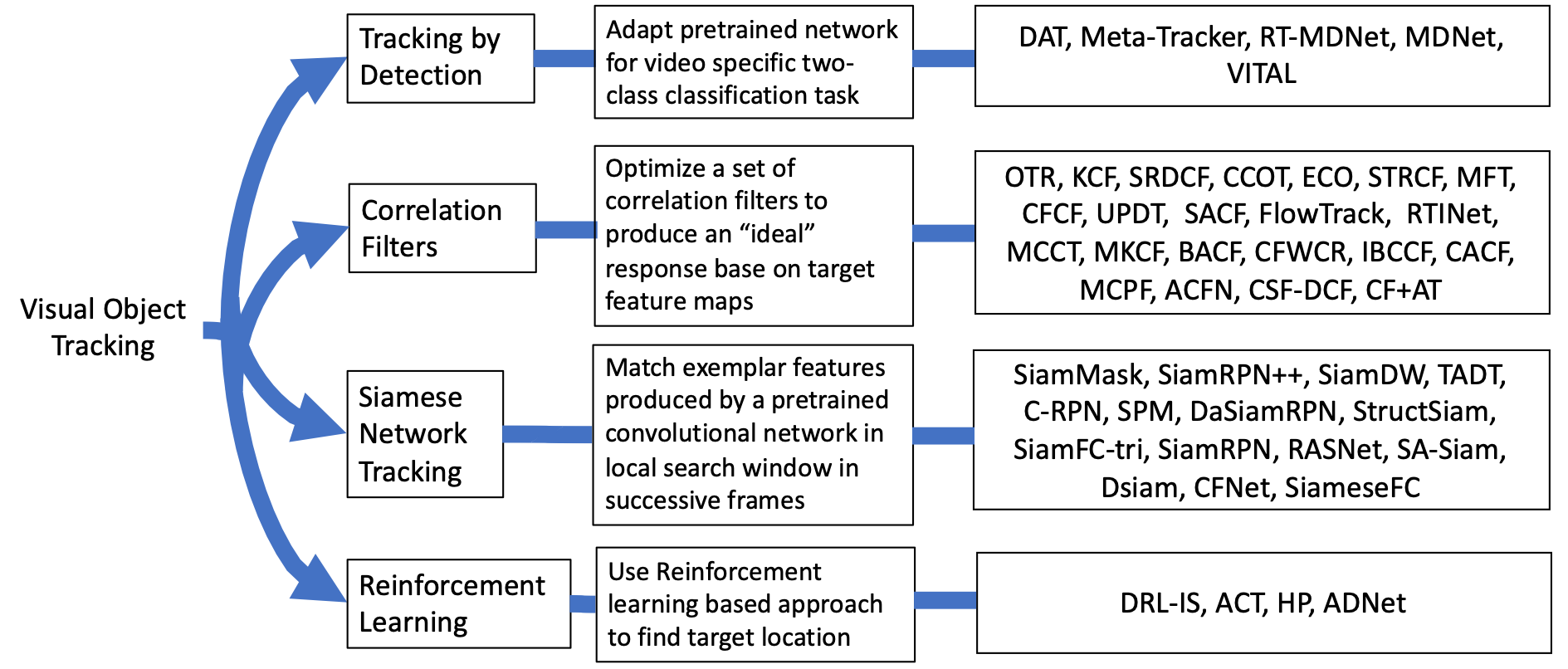}
\end{center}
   \caption{Grouping of convolutional neural network (CNN)-based visual object tracking algorithms.}
\label{fig:long}
\end{figure}

Recent CNN based trackers can be broadly categorized into four groups, as illustrated in Figure~\ref{fig:long}: (i) Tracking by detection (TD), (ii) Correlation Filters (CF), (iii) Siamese networks (SN), and (iv) Reinforcement learning (RL). In CF based trackers, correlation filters are learned to match the target distribution aiming for a response that is Gaussian distributed. The implementation of the CF trackers is usually done in the Fourier domain for computational efficiency and filters are updated during online tracking. However, the filter update reduces the speed of the tracking procedure \cite{danelljan2017eco, danelljan2016beyond, danelljan2015learning}. SN-based trackers approach tracking as a similarity learning problem, where matching is done in feature space. SN trackers are trained offline and are not updated online, which is efficient but may reduce tracking performance \cite{bertinetto2016fully, zhu2018distractor, li2018high}. TD-based trackers treat tracking as a binary classification problem that aims to separate the target from the background. Multiple patches are taken in the target frame ($t^{th}$ frame) near the target location in the previous frame ($(t-1)^{th}$ frame) and the patch with the highest score is selected as the target patch \cite{nam2016learning, jung2018real, shi-nips18-DAT}. RL-based trackers learn an optimal path to the tracked object, either by moving the predicted location to the target location or by learning hyperparameters for tracking \cite{yun2017action, chen2018real}. 

Evaluating tracking algorithms requires large and diverse datasets with annotated ground truth of the target location at every frame. Additionally, attribute annotation is important to fully assess tracker performance in different challenging situations, such as occlusion, illumination variation, etc. 
There are several tracking benchmarks in the literature \cite{WuLimYang13, wu2015object, kristan2018visual, kristan2017visual, kristan2016visual, kristan2015visual, kristan2014visual, kristan2013visual, smeulders2014visual, song2013tracking, li2015nus, liang2015encoding, mueller2016benchmark, li2017visual, kiani2017need, zajc2017beyond, du2018unmanned, muller2018trackingnet, valmadre2018long, fan2018lasot}. 
Two of the most popular tracking benchmarks are the OTB \cite{wu2015object, WuLimYang13} and the VOT challenge \cite{kristan2013visual,kristan2014visual, kristan2015visual, kristan2016visual, kristan2017visual, kristan2018visual} datasets.
In our study we benchmark various trackers on aerial datasets, another important category of datasets that consist of sequences taken from aerial platforms. 
We consider a video to be aerial if the camera is on an airplane or unmanned aerial vehicle (UAV). To include a greater variety of datasets, we broaden this definition to include cases where the camera is located anywhere above the ground ranging from a roof to a satellite. 

There are some key factors that make aerial tracking different from tracking in standard ground level videos. In the aerial videos, the area covered by the field of view of the camera is usually much larger than that of ground level videos. More importantly, in aerial videos, the tracked object is much smaller in size in terms of pixels.
Due to the smaller object size, it is more difficult to generate discriminative features, which adversely affects tracking performance. 
The object's size and viewpoint can change significantly and quickly in aerial videos.
Additionally, the tracked object may be occluded for a long time and even disappear for several frames. 
Camera motion often causes abrupt changes in the object appearance and may result in out-of-frame movement for the tracked object.
These characteristics present a unique set of challenges for aerial tracking compared to standard tracking on ground level videos.
Initial work by Minnehan et al. \cite{minnehan2018benchmarking} indicated that the performance of deep trackers varies significantly when tracking in aerial videos.
Popular trackers, such as tracking by detection MDNet \cite{nam2016learning}, or CF-based CCOT \cite{danelljan2016beyond}, were not as effective when tracking in aerial videos due to occlusion, smaller target size, and camera motion
\cite{minnehan2018benchmarking}. 
Our study differs from \cite{minnehan2018benchmarking} in various aspects. We consider a larger number of more recent trackers from four general groups. We perform our benchmarking on multiple datasets and conduct detailed evaluation for various attributes of aerial imagery.

Several review papers exist in the literature that overview visual object tracking \mbox{research \cite{yilmaz2006object, smeulders2014visual, li2018deep, fiaz2018handcrafted}}. 
Li et al. \cite{li2018deep} studied deep learning trackers based on network architecture, network training, and network function and ran experiments on OTB100, TC-128 \cite{liang2015encoding}, and VOT2015~\cite{kristan2015visual} to compare different deep learning based trackers. 
Fiaz et al. \cite{fiaz2018handcrafted} performed an extensive review that compared various trackers based on different feature extraction methods. Trackers based on both deep learning and hand crafted features were evaluated on benchmarks, such as OTB 2015, OTB 2013~\cite{WuLimYang13}, TC-128, OTTC \cite{fiaz2018handcrafted}, and VOT 2017 \cite{kristan2017visual}. 
VOT challenges compare many different trackers and provide a good overview of the performance of recent trackers.

However, these review papers and tracker benchmark studies deal with datasets that are ground based. 
In this work, we focus on aerial tracking because the conditions encountered vary significantly from ground level situations.
The main contributions of this paper can be described as follows.
\begin{itemize}
    \item We focus on aerial tracking using videos taken from aerial platforms. To our knowledge, this is the first comprehensive benchmarking study of visual object tracking on aerial videos.

    \item We benchmark ten recent deep learning based trackers from four tracking groups. 
    \item We consider four different aerial benchmarks and compare the trackers' performance in various challenging situations which provides a better understanding of the state-of-the-art of visual object tracking on aerial videos.
    

\end{itemize}


The rest of the paper is organized as follows.
In Section \ref{sec:related work}, we discuss the relevant tracking algorithms we incorporated in our benchmarking. 
In Section \ref{sec:experiment}, we discuss the datasets used in our experiments and the evaluation metrics.
In Section \ref{sec:results}, we show the evaluation results on different benchmarks and discuss the comparison among different trackers on different benchmarks as well as specific attributes present within the datasets.
In Section \ref{sec:conclusion}, we present final remarks based on our evaluation.

 \section{Tracking Algorithms}
\label{sec:related work}
In this section, we overview the tracking algorithms based on their approach to tracking and network architectures. We only selected trackers that use deep learning features and considered their performance and source code availability. 

A brief description follows for the four types of trackers outlined in Figure \ref{fig:long}.

\subsection{TD-Based Trackers}
Tracking by detection frameworks view tracking as a foreground (target) vs. background classification problem. TD-based methods have been used for some time and continue to be popular with deep trackers in recent years \cite{nam2016learning,jung2018real, shi-nips18-DAT, song2018vital, park2018meta, fan2017sanet, teng2017robust, nam2016modeling}. In these frameworks, the networks generally learn the region of interest by sampling multiple patches from the input image. Positive and negative instances within the training samples are selected based on the intersection-over-union (IoU) score with the ground truth. 
Online training is usually done in these networks to improve the classification accuracy during tracking.

The TD-based trackers that we included in our benchmarking are MDNet \cite{nam2016learning}, DAT \cite{shi-nips18-DAT}, Meta-Tracker \cite{park2018meta}, and RT-MDNet \cite{jung2018real}. MDNet was the winner of the VOT2015 challenge and was used as a baseline tracker for the other TD-based trackers. DAT improved over MDNet using reciprocation learning. The Meta-Tracker improved over MDNet using a meta-learning approach and RT-MDNet improved the real time performance of MDNet. Short descriptions of these trackers are given below.

\subsubsection{Multidomain Network Tracker (MDNet)}
In visual object tracking,
there are some desirable properties for target representation learning such as invariance with respect to illumination, scale, perspective, and motion blur. The goal of using multidomain learning is to learn a discriminative model that learns a shared representation of the target in various domains  \cite{nam2016learning}. To achieve this, MDNet \cite{nam2016learning} is trained offline with large set of video sequences, where each sequence is considered as a domain. 

\begin{figure}[H]
\begin{center}
   \includegraphics[width= 0.8\textwidth]{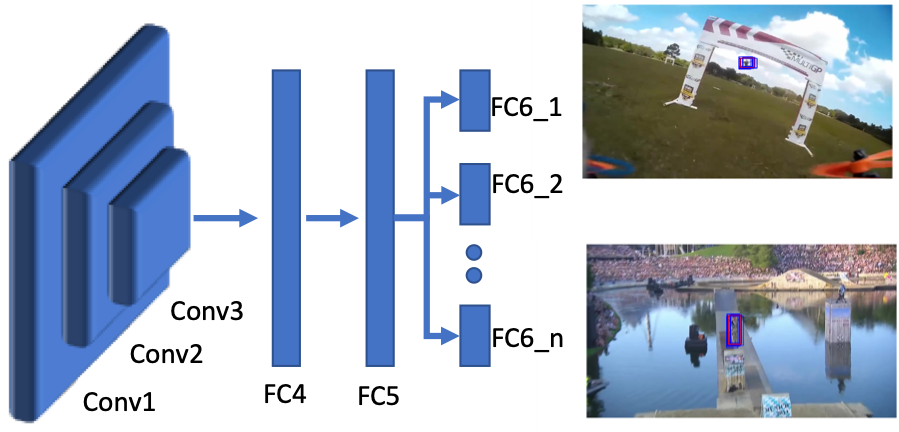}
\end{center}
   \caption{Network architecture of the MDNet tracker \cite{nam2016learning}.}
\label{fig:mdnet}
\end{figure}

The MDNet \cite{nam2016learning} architecture is shown in Figure \ref{fig:mdnet}. 
In the network shown, the domain specific layers are shown as FC6\_1 to FC6\_n where all the preceding layers are considered as shared layers.
During tracking, all branches of the sixth fully connected layer are removed and replaced with a single fully connected layer, where online adaptation is performed by fine-tuning the fully connected layers.
For~precise localization of the object during tracking, a bounding box regression technique is used on bounding boxes with high scores.  

\subsubsection{Deep Attentive Tracking (DAT)}
DAT \cite{shi-nips18-DAT} learns to attend a specific foreground class from the background by using reciprocation learning.
Here, reciprocation learning refers to utilizing the backpropagated gradient to aid the learning procedure. It is achieved by passing the input image through the network and gathering the backpropagated gradients in the input layer which are later used as a regularization term with the classification loss.
The MDNet architecture is used for the implementation. However, VOT benchmark training is no longer required for this tracker. Instead, for the first three convolutional blocks, VGG-M weights pretrained on ImageNet \cite{ILSVRC15} are used and never updated. 
Similarly to MDNet, the fully connected layers are updated during online fine tuning. 
During this process, multiple patches are sampled at every input frame. 
The sampled patches are passed through the network and an attention map is obtained by taking the partial derivative of the classification score with respect to the input image at each patch. 
This attention map regularizes the original loss function through an additional term, with the goal of better classification where the map localizes the target.
The objective is to maximize the mean and minimize the standard deviation of the attention map corresponding to the true class and do the inverse for the background. 
Eventually a hyperparameter is added to combine the two losses. Evaluation on OTB benchmarks showed the effectiveness of this tracking algorithm. 

\subsubsection{Meta-Tracker (Meta-SDNet)}
Meta-SDNet \cite{park2018meta} improves over the baseline MDNet tracker using a meta-learning approach
for online adaptation taking into consideration the uncertainty during tracking. 
Meta learning refers to a few shot learning procedure, where an algorithm adapts in a new environment by learning either model parameters, or a metric, or proper optimization techniques.
In the meta-learning process, the network parameters are learned such that uncertainties in future frames can be minimized by better modeling the target appearance without overfitting the recent target appearances. 
Specific video sequences suitable for training were selected from a large scale video detection dataset and used to learn the appropriate parameters for meta-learning. 
The first three layers of Meta-SDNet are based on the pretrained VGG16 framework and used as feature extractors. The last three fully connected layers were randomly initialized and trained with the Adam optimizer.  

\subsubsection{Real-Time MDNet (RT-MDNet)}
RT-MDNet \cite{jung2018real} improves over MDNet tracking framework by incorporating ROI alignment technique from Fast-RCNN \cite{girshick2015fast}. The technique allowed to construct a high resolution feature map.
The channel activation was based on large receptive field.
An adaptive ROI alignment layer was added after the convolutional layers and before the fully connected layers in the original MDNet architecture. 
This~feature extraction method improves the computational complexity of the overall tracking process. 
Dilated convolution was used to extract high resolution feature maps, which improved the quality of the representation of the target in feature space. 
Modified bilinear interpolation was considered for the adaptive ROIAlign layer instead of linear interpolation which changed the tracking performance significantly. 
Another contribution was combining instance embedding loss with the classification loss to discriminate between similar foreground targets in multiple domains. 
The network was tested on OTB 2015 and UAV123 datasets \cite{mueller2016benchmark} and performed comparable to the state-of-the-art trackers in real-time.

 \subsection{CF-Based Trackers}
\begin{figure}[H]
\begin{center}
   \includegraphics[width=0.9 \textwidth]{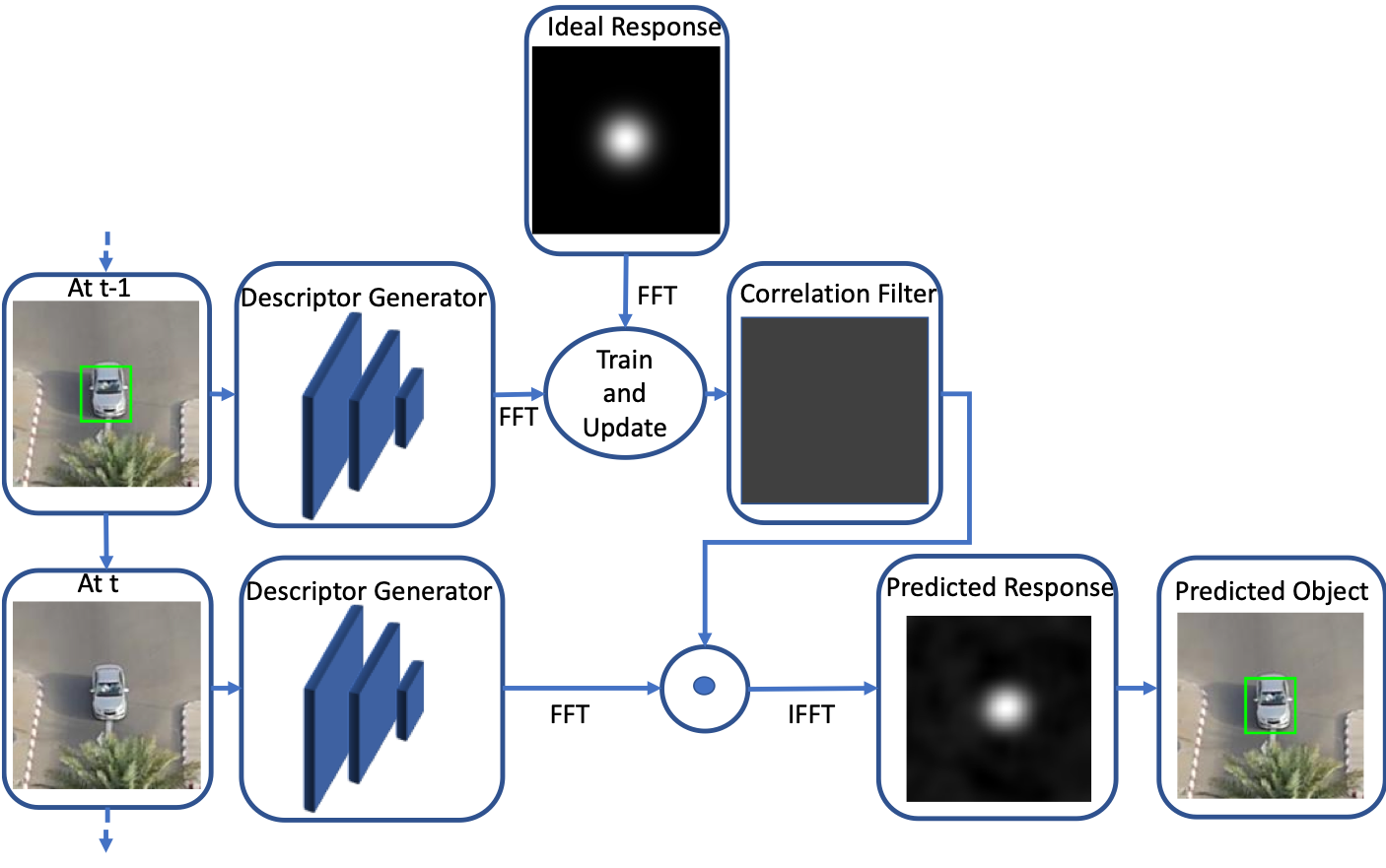}
\end{center}
   \caption{A general deep feature-based correlation filter tracking framework.}
\label{fig:cf}
\end{figure}

A popular method for visual object tracking is learning Discriminative Correlation Filters (DCF) to predict the location of the tracked object in a patch \cite{valmadre2017end, kart2018object, gundogdu2018good, bhat2018unveiling, zhang2018visual, zhu2018end, yao2018joint, wang2018multi, tang2018high, he2017correlation, li2017integrating, mueller2017context, zhang2017multi, choi2017attentional, bibi2016target}. 
A basic correlation filter based tracking framework is shown in Figure \ref{fig:cf}.
Generally, a large patch around the tracked object is cropped at $t^{th}$ frame during tracking. Any feature extraction technique may be used to extract features from the cropped patch. 
Then the features are utilized to learn a bank of correlation filters that generates a Gaussian response map at the desired target location.
Based on the response map, the bounding box of the tracked object is predicted.  
However, the filter learning procedure is performed at various time instances. Generally, a few frames and the corresponding target locations are saved based on the tracking confidence and utilized during the filter learning procedure.

CF-based trackers aim to find the filters $f$ where the template $x$ is given from a patch with ideal response map $y$, typically modeled by a Gaussian distribution. For example, the Kernelized Correlation Filter (KCF) \cite{henriques2015high}  utilized a Gaussian kernel to model the target response. The best filter parameters are computed as follows.
 
\begin{equation} \label{eq:cf}
     f^{*} = \argmin_{f\in \R^{m \times m}} \norm{\sum_{c=1}^{D} x^c \star {f^c} - y }^{2}_2
\end{equation}

 
However, CFs have limited detection range and may perform poorly when the object undergoes deformation. The patch size and the filter size must be equal, which often makes the tracker learn the background within the patch for irregularly shaped objects. If the patch is small, then in cases of occlusion the object may not be redetected after reappearing. Therefore, it is important to incorporate regularization in the CF-based tracking framework. 

Danelljan et al. proposed Spatial Regularization for learning the DCFs (SRDCF \cite{danelljan2015learning}). The~objective was to weaken the responses due the background information by spatially modifying the filter coefficients. The background is suppressed by assigning higher values of the filter coefficients which are outside of the target bounding box and vice versa. The filter parameters $f^*$ are learned using Equation (\ref{eq:srdcf}) where $\alpha_t$ describes the impact of each training sample and $w$ is the spatial regularization term.  
\begin{equation}\label{eq:srdcf}
     f^{*} = \argmin_{f\in \R^{m \times m}} \sum_{t=1}^{T} \alpha_t \norm{\sum_{c=1}^{D} {x_t}^c \star {f^c} - y_t }^{2}_2 + \sum_{c = 1}^{D}\norm{w \cdot f^c}^2_2
\end{equation}

An improved version of SRDCF is the Continuous Convolution Operation Tracker (CCOT \cite{danelljan2016beyond}), where filters are learned for multiple resolutions of target patches in the continuous domain. These filters are then used to produce multiple resolution feature maps. However, CCOT suffers from the large number of filters that are required to be learned to capture the target representation. Another limitation is that the tracker updates at every frame, which causes overfitting to the most recent target appearance. 


There are many variations and improvements to CF trackers since SRDCF and CCOT.
We selected the following CF based trackers in our benchmarking.
The Efficient Convolution Operators (ECO)~\cite{danelljan2017eco}, the Spatial Temporal Regularized Correlation Filter (STRCF) \cite{li2018learning}, and the Multi-Fusion Tracker (MFT)~\cite{bai2018multi}. 
ECO was the winner tracker of VOT2016 challenge. 
MFT was the winner tracker of VOT2018 challenge. 
STRCF is also one of the top performing CF-based trackers. 
A brief overview of these approaches is given below.  

\subsubsection{Efficient Convolution Operators (ECO)}
The ECO \cite{danelljan2017eco} formulation is based on discriminative correlation filters 
with a factorized convolution operator introduced to reduce overfitting, a generative model to estimate the training sample distribution, and an efficient model update strategy. 
The base framework of CCOT had many filters that contained negligible energy, and these were eliminated in ECO to make the training more efficient. Then, filters are reproduced as the linear combination of the learned filters. 
The Gauss--Newton method was used to optimize the quadratic loss using conjugate gradient method. 
The factorized convolution operation reduces the computational and memory complexity of the tracker. 
To improve overfitting compared with CCOT, a new sample space was introduced based on Gaussian mixtures to obtain a representative sample set. A model update strategy was also introduced to reduce overfitting based on updating the model every $N_s$ frames, where this parameter was identified by heuristics. 
Small values of $N_s$ may result in overfitting, whereas large values reduce the convergence speed of the optimization. The base network model was VGG, where features from the first and last convolutional layers were used along with HOG and Color features.
Finally, the comparison was done on the tracking benchmarks and the model achieved state-of-the-art performance on the VOT2016 challenge.     

\subsubsection{Multi-Fusion Tracker (MFT)}
The Multi-Fusion Tracker (MFT) \cite{bai2018multi} improves upon the baseline SRDCF tracker by using a motion model and hierarchical feature selection with adaptive fusion.  
Typically, motion models are ignored in CF and TD trackers. However, MFT utilized a motion model to improve over partial occlusion and bounding box estimation. For the motion model, Kalman filtering was used, which effectively reduces the noise of the bounding box center locations from the DCF predictions.
Another important distinction is that the online training was formulated to learn independent correlation filters for multilevel CNN features. It was determined that early layer features are better for adapting the scale changes for small deformation, but deeper features are better for adapting with larger deformation.
Additionally, middle-level features are the most representative of the target scale, as the deeper features are prone to drifting towards similar objects. 
This shortcoming is solved by fusing multilevel features from the network. For these multilevel features, adaptive independent correlation filters were learned using conjugate gradient method. The model was updated at a fixed frame interval instead of every frame following ECO. Then, the outputs of the multiple independent hierarchical filters are fused using an adaptive weighting scheme, where the center location of the target bounding box is extracted from the feature map. 
Scale change is achieved using a multiscale search strategy of the image patches, after~cropping based on the motion estimation model which predicts the center location of the patch. The MFT algorithm outperforms the baseline ECO architecture on the OTB dataset.

\subsubsection{ Spatial Temporal Regularized Correlation Filter (STRCF)}
Spatial Temporal Regularized Correlation Filter (STRCF) \cite{li2018learning} tracking incorporates both spatial and temporal regularization on the DCF framework. In a comparative study on different sequences where the target appearance varies significantly, STRCF outperforms SRDCF due to its effective appearance modeling. To solve for the filter parameters, the Alternating Direction Method of Multipliers (ADMM) was used to achieve closed form solutions. The algorithm can operate in real-time when using handcrafted features. With the temporal regularization term, the objective function to update the filters $f^*$is given by
\begin{equation} \label{eq:strcf}
     f^{*} = \argmin_{f\in \R^{m \times m}} \frac{1}{2} \norm{\sum_{c=1}^{D} {x_t}^c \star {f^c} - y_t }^{2}_2 + \frac{1}{2}\sum_{c=1}^{D}\norm{w \cdot f^c}^2_2 + \frac{\mu}{2}\norm{f - f_{t-1}}^2_2
\end{equation}

Here, $f_{t-1}$ represents the filters used in the $(t-1)^{th}$ frame and $\mu$ denotes the hyperparameter for regularization. This expansion to online passive-aggressive algorithm \cite{crammer2006online} improves over SRDCF in two ways: (i) better model updating with multiple samples and (ii) better occlusion handling by passively updating the correlation filters. Evaluation results on Temple-color, VOT-2016, and OTB 2015 benchmarks showed that this algorithm achieved state-of-the-art accuracy. 

\subsection{SN-Based Trackers}
Siamese networks contain twin branches with shared weights and are widely used in visual object tracking \cite{bertinetto2016fully, Wang2019SiamMask, li2018siamrpn++, SiamDW_2019_CVPR, TADT, fan2018siamese, wang2019spm, zhang2018structured, dong2018triplet, wang2018learning, guo2017learning, minnehan2019fully}.
The objective of Siamese networks is to learn a shared representation of the input images in a similarity learning fashion. Generally, two input images are fed to the network in each of the branches, and both branches of the network are updated at the same time with shared weights.
Bertinetto and Valmadre et al. \cite{bertinetto2016fully} proposed a fully convolutional Siamese architecture for tracking (SiamFC) that has a template branch and a search window branch. The network was trained with the ImageNet video detection dataset \cite{ILSVRC15}, where two different frames in a sequence are cropped and resized such that  
the area $A$ of the resized patch is
\begin{equation}
 A = s(w+2p) \cdot s(h+2p)
\end{equation}
where $w$ and $h$ are the width and height of the corresponding target bounding box, $p$ is the context amount which is set to $p = (w+h)/4$, and $s$ is the scale factor. 
Finally, cross-correlation is applied in feature space to get the final response map.
Multiples scales and aspects are considered to deal with the scale and aspect ratio changes. 
Training the network is done using positive and negative pairs with logistic loss 
\begin{equation}
l(y,v) = log(1 + exp(-yv))
\end{equation}
where $v$ is the score for one pair of patches and $y\in\{ +1,-1\}$ is the target value. 
The overall network is trained using SGD with average loss.

One limitation of SiamFC is that it does not always find the tightest bounding box around the target and the resulting localization accuracy is not as good compared to the CF-based trackers. \mbox{Li et al. \cite{li2018high}} incorporated the Faster R-CNN \cite{ren2015faster} with the SiamFC architecture. The bounding box proposal generation and bounding box regression from Faster R-CNN improved the overall performance of the tracking framework. He et al. \cite{he2018twofold} proposed a twofold Siamese network architecture where semantic appearance information is encoded to get a better response map. Zhu et al. \cite{zhu2018distractor} further improved SiamRPN using distractor-aware training. They also used a local to global search strategy to improve tracking during occlusion. 

Among SN-based trackers, we select SiamFC \cite{bertinetto2016fully} and DaSiamRPN \cite{zhu2018distractor}. SiamFC was the winning tracker of the VOT2017 realtime challenge and DaSiamRPN was the winning tracker of the VOT2018 realtime challenge. Furthermore, these trackers are the backbone of many other state-of-the-art SN-based trackers \cite{Wang2019SiamMask, li2018siamrpn++, SiamDW_2019_CVPR, TADT}. Brief descriptions about these trackers are provided below.   

\subsubsection{ Fully Convolutional Siamese Tracker (SiamFC)}

\begin{figure}[H]
\begin{center}
   \includegraphics[width=0.8 \textwidth]{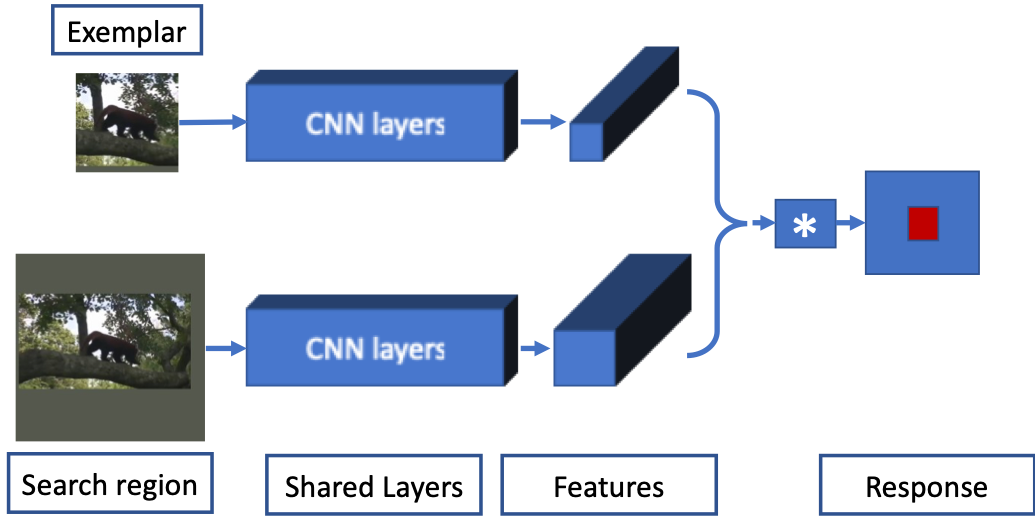}
\end{center}
   \caption{SiamFC network architecture. \cite{bertinetto2016fully}}
\label{fig:siamese}
\end{figure}

SiamFC \cite{bertinetto2016fully} casts the tracking problem as a similarity learning problem and uses fully convolutional branches for feature embedding. 
The SiamFC network architecture is shown in Figure \ref{fig:siamese}. 
Two input images of different size are fed into the two branches of the network and the same transformation is applied to both images.
Keeping the target at the center, the first frame of the sequence is cropped and resized to pass through one of the branches of the Siamese network. 
The first frame of the sequence is kept fixed in one of the branches in the network. 
All the other frames of the sequence are cropped, resized, and passed through the second branch of the network. 
After getting the feature embedding from both branches, a correlation layer is used to find the correlation between two images in the latent space. In the final correlation map, the bounding box is determined in the detection frame based on the similarity score. 
In the SiamFC architecture, binary cross-entropy loss with SGD is used during training. The network is trained offline on the 2015 version of ImageNet Large Scale Visual Tracking Challenge \cite{ILSVRC15} dataset and no online update is made during tracking. One drawback of this approach is that it uses an expensive multiscale test to adapt for changes in the object's scale, which is not very efficient and does not capture the scale change very well.  

\subsubsection{Distractor-Aware Siamese Region Proposal Networks (DaSiamRPN)}
The initial SiamRPN \cite{li2018high} architecture utilizes a key concept in Siamese networks, where the goal is to learn an embedding space maximizes the distance between the inter-class objects and minimizes the distance between intra-class objects. SiamRPN introduces several new concepts including a Region Proposal Network (RPN) and one-shot learning. 
The region proposal network is utilized on top of the base Siamese network adopted from the original SiamFC architecture for feature extraction. 
The~DaSiamRPN \cite{zhu2018distractor} network improves the training procedure by increasing the number of positive examples to reduce the imbalance between positive and negative examples. The original SiamRPN is trained on Youtube-BB dataset \cite{real2017youtube} consisting of 200,000 video sequences which are annotated every 30 frames. In DaSiamRPN, the ImageNet Detection \cite{ILSVRC15} and COCO Detection \cite{lin2014microsoft} datasets were augmented into image pairs, and used to train the network. 
Semantic negative pairs were included, instead of considering hard negative pairs as in object detection. Additionally, motion blur images were used for training. 

Distractors were collected using non maximum suppression to avoid redundant candidates. 
Among all the predicted bounding boxes, the box with highest score was set as the target and boxes with score greater than a threshold were chosen as distractors. 
This approach works well for short-term tracking, but it may fail for long-term tracking, as the search region is not large enough to cover the entire range within the image where the object may reappear. DaSiamRPN tried to solve this issue by using local-to-global search strategy when tracking failure occurs and keeping track of the number of frames in which the target is not found. 
For the training architecture, a modified AlexNet pretrained on ImageNet was used. Evaluation on multiple benchmarks showed that the tracker achieves state-of-the-art performance. 

\subsection{RL-Based Trackers}

In RL-based approaches an agent learns to find an optimal path in an environment from its own experience using feedback. 
The agent generally receives observations in discrete time steps with rewards and chooses an action from a set of available options. This process continues until convergence. Reinforcement learning algorithms are extensively used in game theory and have been considered for visual object tracking as well \cite{ren2018deep, chen2018real, dong2018hyperparameter, yun2017action, supancic2017tracking, huang2017learning}. 
For example, in the HP \cite{dong2018hyperparameter} tracker, Dong et al. introduced a novel hyper parameter selection technique which can learn sequence specific hyperparameters using continuous deep Q-learning. 
Supancic et al. \cite{supancic2017tracking} formulated the tracking problem as an online partially-observable Markov decision making process (POMDP) where, instead of heuristic target initialization, an optimal decision making policy is learned to update the appearance model. Among the RL-based trackers, the Actor-Critic \cite{chen2018real} tracker outperformed the other trackers based on the evaluation results provided in the corresponding papers. 
We have included the Actor--Critic tracker in our benchmarking and provide a short description below. 
\subsubsection{Actor--Critic Tracking (ACT)}
ACT \cite{chen2018real} is a reinforcement learning tracker that can operate in real-time. 
The two main components of the framework are the Actor and the Critic models. 
The actor network moves the bounding box to the target location and the critic network guides the learning process during offline training.
The~process is guided by calculating the Q-value using reinforcement learning to train both the Actor and the Critic networks. 
A modified deep deterministic policy gradient algorithm is used to effectively train the model. 
During online tracking, the Actor model employs a dynamic search framework to learn the position of the target and the Critic model verifies the position to make the tracker more robust. 
A~pretrained VGG architecture was used to initialize the Actor and Critic networks. 
In the Critic network, Q learning was done using the Bellman equation in Q-learning, while the Actor network learns using chain rule. 
During training, the samples were generated from translation and scaling of the bounding box, where the scale was sampled from a Gaussian distribution centered by the object location. 
The ImageNet Videos were used to train the Actor network so that for each iteration 20 to 40 frames were randomly chosen for training. 
The tracker achieves 30 fps speed with performance comparable  with state-of-the-art trackers on popular tracking benchmarks.

\section{Experiments}
\label{sec:experiment}

In our experiments, we evaluated a subset of the OTB 2015 \cite{wu2015object} dataset, the DTB70 \cite{li2017visual} dataset, the UAV123 \cite{mueller2016benchmark} dataset and the UAV20L \cite{mueller2016benchmark} dataset as our benchmark datasets. For the aerial style subset of OTB 2015 dataset, we selected the sequences where the camera  is above the ground. The chosen sequences are: Basketball, Bolt, CarScale, Couple, Crossing,   Crowds, Human3, Human4-2, Human5, Human6, Jogging-1, RedTeam, Subway, SUV, Walking, Walking2, and Woman.

The DTB70 dataset contains 70 sequences of UAV collected data where the bounding boxes are drawn manually. Some of the sequences are collected from YouTube. 
Different types of camera motion, including translation and rotation, are incorporated to make the dataset more challenging. 
Three types of targets appear in the videos: human, animal, and rigid objects. 

The UAV123 dataset contains 123 video sequences taken from UAV platforms. 
Note that we excluded the seven synthetic sequences from the UAV123 dataset for our evaluation. 
A subset of the UAV123 dataset, UAV20L, is also evaluated for long-term tracking analysis. 

The attributes of the aerial datasets are listed in Table \ref{tab:attributes}. These attributes make aerial tracking more challenging and their annotations are available within the corresponding datasets.  
The comparison of different trackers with these attributes provide better understanding of their performance under different tracking scenarios. Comparisons across UAV123 and DTB70 will indicate the generalizability of different trackers. For long-term tracking, an important consideration is consistent performance in a long temporal span, which tests the tracker's ability to create a robust model and perform efficient model updates. 
Some trackers may drift a little from one frame to the next, which may not be noticeable in short term sequences, but eventually could result in target loss during long-term tracking. 

Regarding the datasets, there are inherent differences between the OTB aerial subset and other aerial datasets in terms of resolution, attributes, and size of the objects to be tracked. 
In the standard OTB sequences, the objects are much larger and occupy a larger portion of the image frame, while in aerial datasets such as DTB70, UAV123, and UAV20L, the object to track takes a smaller portion of the image because the camera is higher and covers a larger area. 
Another important distinction of the DTB or UAV sequences is that the camera rotates around the object, whereas none of the videos in OTB sequences have this attribute. 
Additionally,  the OTB benchmark sequences have minimal camera jitter and less clutter in the background. 

\begin{table}[H]
\caption{Dataset attributes.} 
\label{tab:attributes}
\centering
{\begin{tabular}{cccccc} 
\toprule
\multicolumn{3}{c}{\textbf{UAV123}} & \multicolumn{3}{c}{\textbf{DTB70}} \\
\midrule
\textbf{Abb} & \textbf{Full Name} & \textbf{Total Sequence} & \textbf{Abb} & \textbf{Full Name} & \textbf{Total Sequence} \\
\midrule
ARC & Aspect Ratio Change & 65 & ARV & Aspect Ratio Variation & 25 \\
BC & Background Clutter & 21 & BC & Background Clutter & 13 \\
CM & Camera Motion & 64 & DEF & Deformation & 18 \\
FM & Fast Motion & 22 & FCM & Fast Camera Motion & 41 \\
FOC & Full Occlusion & 33 &  IPR & In-plane-rotation & 47\\
IV & Illumination Variation & 25 & MB & Motion Blur & 27 \\
LR & Low Resolution & 48 & OPR & Out-of-plane Rotation & 06 \\
OV & Out of View & 28 & OV & Out-of-view & 07 \\
POC & Partial Occlusion & 68 &  OCC & Occlusion & 17 \\
SOB & Similar Object & 39 & SOA & Similar Object Around & 27 \\
SV & Scale Variation & 103 & SV & Scale Variation & 22 \\
VC & Viewpoint Change & 55 &  & & \\
\bottomrule
\end{tabular}}
\end{table}

Tracker codes were obtained from their Github repository at the URLs provided in Table \ref{tab:resources}. None of the algorithms in our implementation were trained from scratch. 
All of the trackers were implemented in a server workstation with NVIDIA TITAN-V GPU. The CF-based algorithms were implemented using MATLAB and MATConvNet, whereas the other trackers are implemented using Python and PyTorch. 
Readers are referred to the Github pages in Table \ref{tab:resources} for further implementation details. 

All the trackers were evaluated using one pass evaluation (OPE) introduced by the OTB benchmark evaluation procedure. The trackers were initialized in the very first frame and never re-initialized after a tracking failure. All the trackers were set to provide the output in the OTB format $([tclx, tcly, w, h])$, where $tclx$ and $tcly$ are the coordinates of the top left corner of the bounding box, respectively, and $w$ and $h$ are the width and height of the box, respectively. 

To evaluate tracker performance on the aerial benchmarking datasets, we examined visual examples as well as results based on evaluation metrics. 
To assess overlap performance, successful tracking is considered if the predicted bounding box and the groundtruth bounding box have an intersection over union (IOU) overlap greater than or equal to some threshold (e.g., 0.5). 
The tracker is evaluated for different thresholds and the success vs. threshold plot is obtained. 
The area under the curve (AUC) is computed based on the success vs. threshold plot and the trackers are ranked based on this value.
\begin{table}[H]
\caption{URLs for the codes of the implemented trackers. The code is denoted as P when the implementation is in Python and PyTorch and M when the implementation is in MATLAB and MatConvNet. MDNet is denoted by py-MDNet since it is based on a PyTorch implementation.} 
\label{tab:resources}
\centering       
{\begin{tabular}{cccc} 
\toprule
\textbf{Tracker} & \textbf{Base Network} & \textbf{Code} & \textbf{Code Repository} \\
\midrule
ACT & MDNet & P & \url{https://github.com/bychen515/ACT}  \\
DaSiamRPN & SiamFC & P & \url{https://github.com/foolwood/DaSiamRPN}  \\
DAT & MDNet & P & \url{https://github.com/shipubupt/NIPS2018} \\
ECO & SRDCF & M & \url{https://github.com/martin-danelljan/ECO} \\
meta-SDNet & MDNet & P & \url{https://github.com/silverbottlep/meta\_trackers} \\

MFT & SRDCF & M & \url{https://github.com/ShuaiBai623/MFT} \\
py-MDNet & MDNet & P & \url{https://github.com/HyeonseobNam/py-MDNet} \\
RT-MDNet & MDNet & P & \url{https://github.com/IlchaeJung/RT-MDNet} \\
SiamFC & SiamFC & P & \url{https://github.com/HengLan/SiamFC-PyTorch} \\
STRCF & SRDCF & M & \url{https://github.com/lifeng9472/STRCF} \\
\bottomrule
\end{tabular}}

\end{table}


\section{Results and Discussion}
\label{sec:results}
\begin{figure}[H]
\begin{subfigure}{.5\textwidth}
  \centering
  \includegraphics[width = \textwidth]{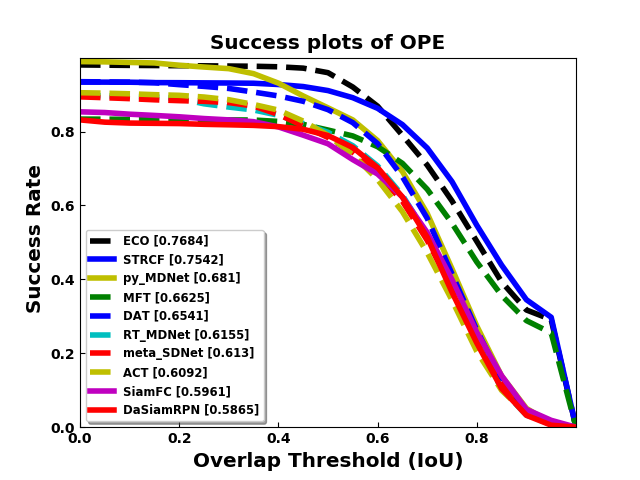}
  \caption{OTB aerial}
\end{subfigure}
\begin{subfigure}{.5\textwidth}
  \centering
  \includegraphics[width = \textwidth]{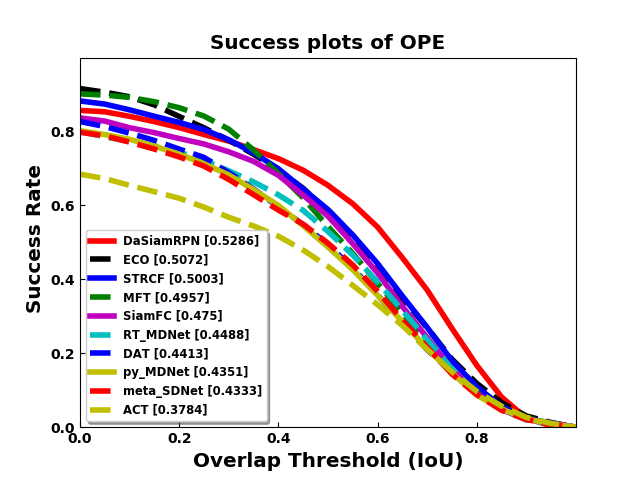}
  \caption{DTB70}
\end{subfigure} \\
\par\bigskip
\begin{subfigure}{.5\textwidth}
  \centering
  \includegraphics[width = \textwidth]{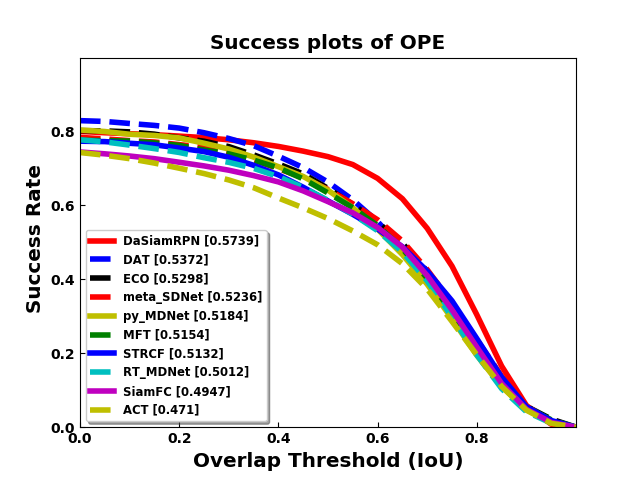}
  \caption{UAV123}
\end{subfigure}%
\begin{subfigure}{.5\textwidth}
  \centering
  \includegraphics[width = \textwidth]{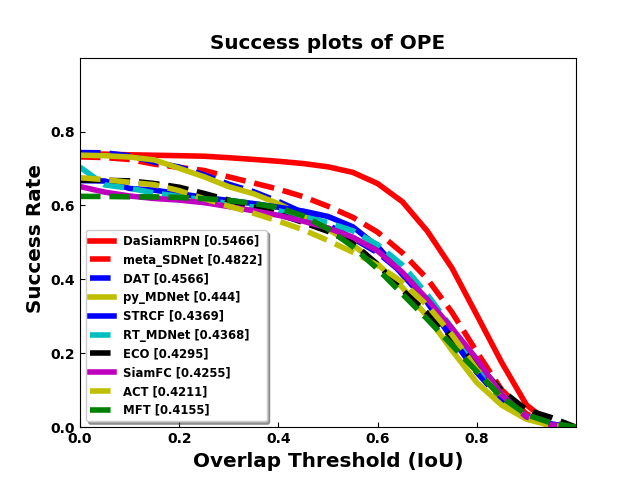}
  \caption{UAV20L}
\end{subfigure}
\caption{Overlap success plots of OPE for the aerial datasets. Results show that DaSiamRPN outperforms all other trackers in DTB70, UAV123, and UAV20L datasets. ECO performed well in the OTB aerial subset. Best viewed zoomed in and in color.}
\label{fig:overallsuccess}
\end{figure}

\begin{figure}[H]
\centering
  \begin{tabular}{@{}llll@{}}
    \includegraphics[width=.32\textwidth]{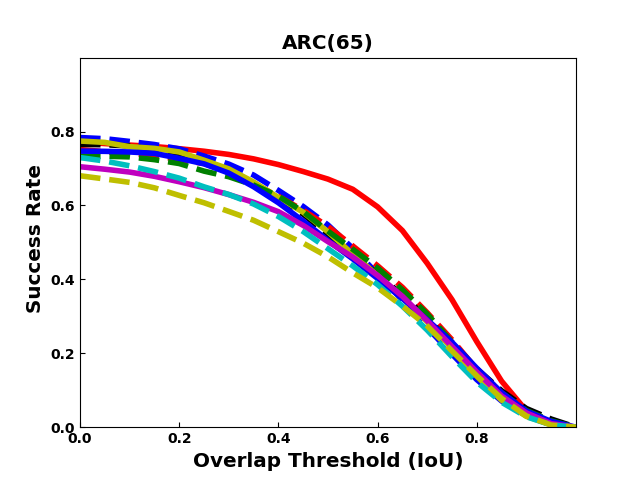} &
    \includegraphics[width=.32\textwidth]{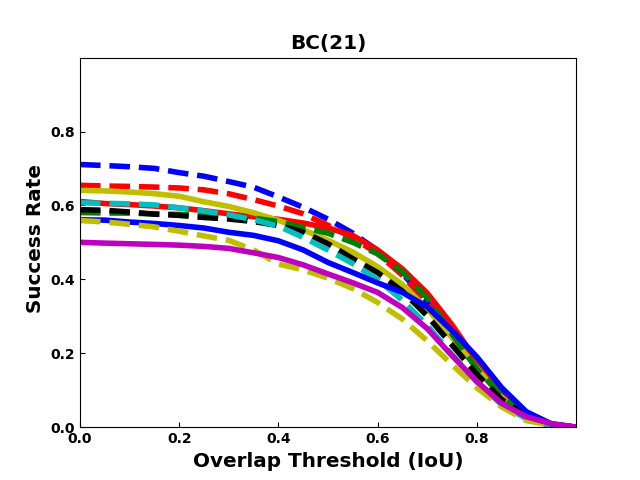} &
    \includegraphics[width=.32\textwidth]{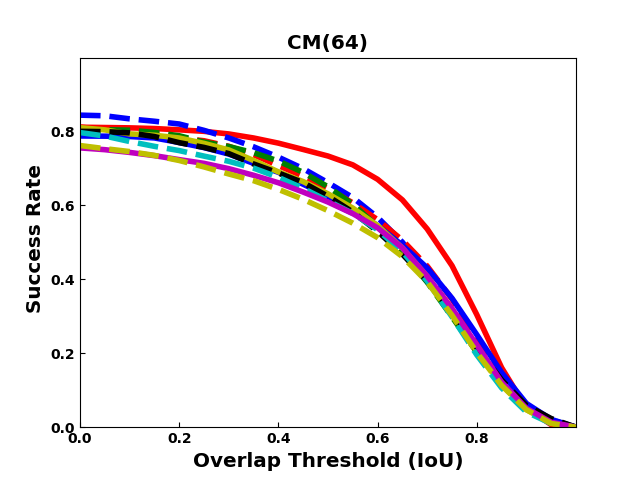} \\
    \includegraphics[width=.32\textwidth]{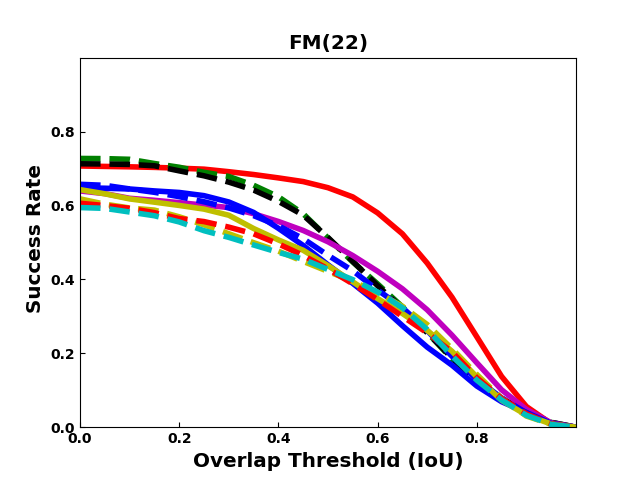} &
    \includegraphics[width=.32\textwidth]{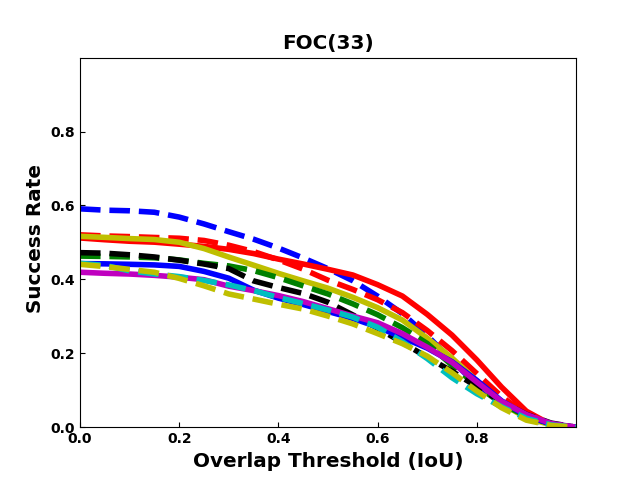} &
    \includegraphics[width=.32\textwidth]{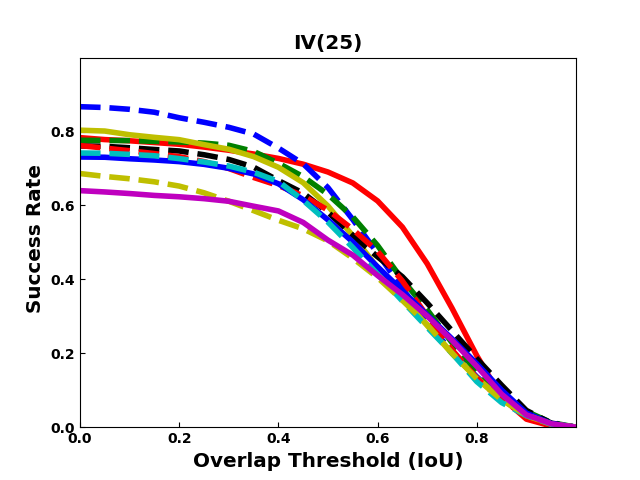} \\
    \includegraphics[width=.32\textwidth]{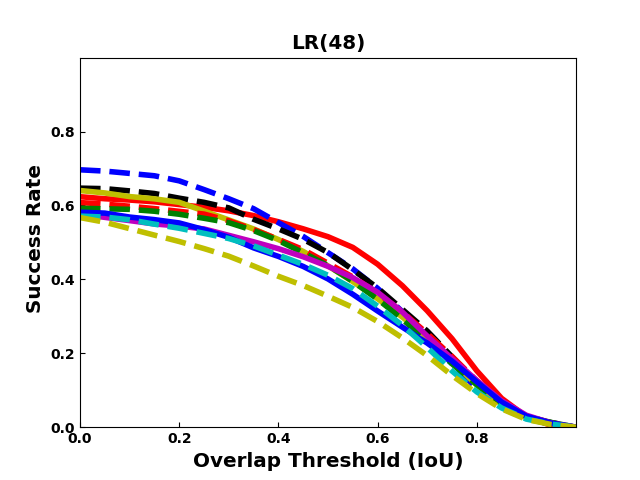} &
    \includegraphics[width=.32\textwidth]{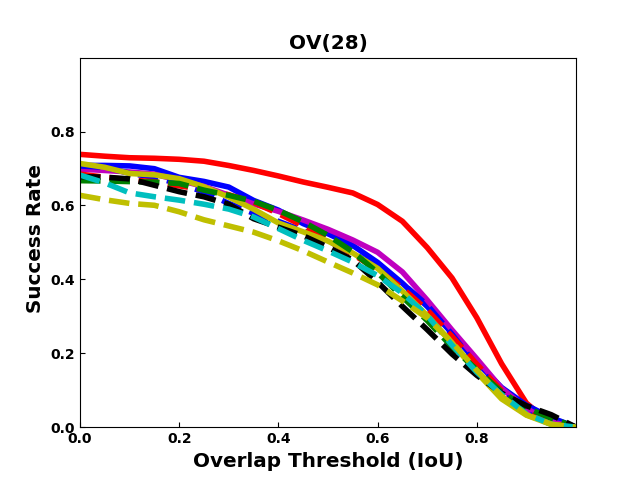}  &
    \includegraphics[width=.32\textwidth]{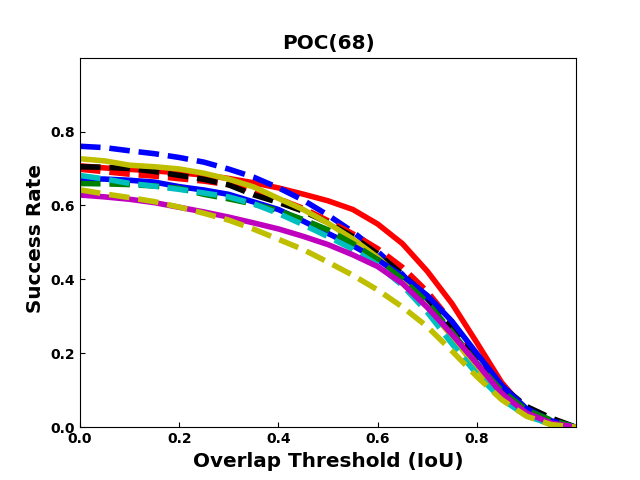} \\
    \includegraphics[width=.32\textwidth]{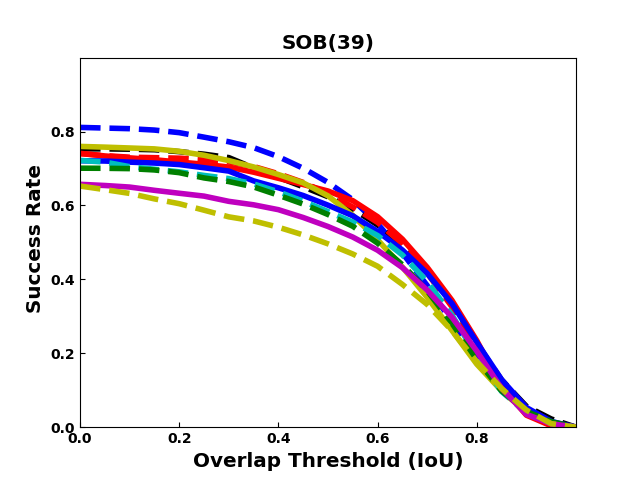} &
    \includegraphics[width=.32\textwidth]{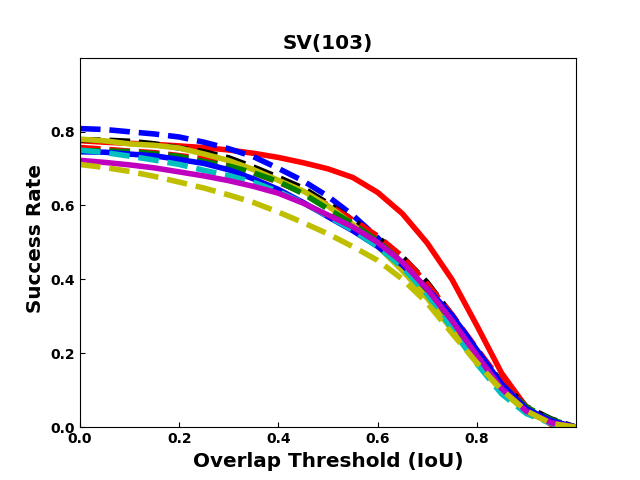} &
    \includegraphics[width=.32\textwidth]{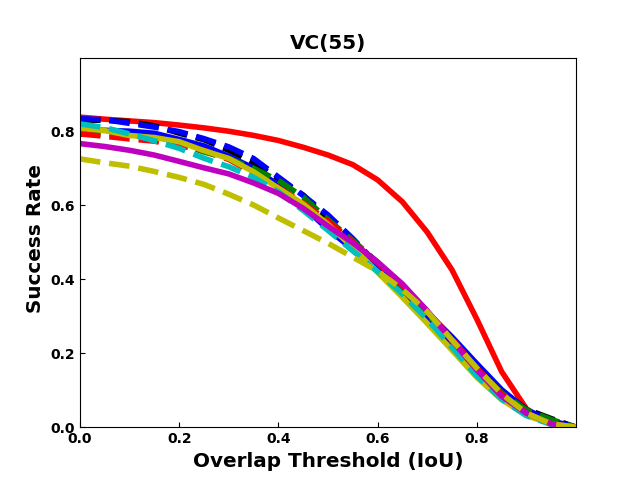}   \\
  \end{tabular}
  \includegraphics[width=\textwidth]{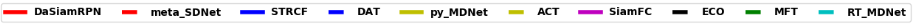} 
  \caption{Overlap plots of OPE for UAV123 dataset attributes. DaSiamRPN outperforms other trackers in most of the challenges except BC, FOC, and SOB. Best viewed in color.}
  \label{fig:uavsuccess}
\end{figure}

In this section, we have present our benchmarking results.
In Figure \ref{fig:overallsuccess}, the overlap success plot is shown for all benchmark datasets. The AUC, computed for each of the trackers, is indicated in brackets in the legends. The results show that DaSiamRPN outperforms the other trackers in three out of four datasets. In Figure \ref{fig:uavsuccess}, the overlap success plot is shown for various attributes in the UAV123 dataset. The AUC is computed and shown in the legends. This plot shows how well various trackers perform in specific challenges such as occlusion, out of view, fast motion, etc. It can be seen from these results that DaSiamRPN does the best in most of the challenges. 
In Table \ref{tab:overallcomparison}, tracker performance (in terms of AUC) is compared for the ground based OTB dataset and the aerial datasets. The results show that even though the AUC for all trackers is high for OTB datasets, their performance was significantly reduced in the aerial datasets.  

\unskip

In terms of precision, the success rate of each tracker is evaluated based on the center to center distance, between the predicted bounding box and the groundtruth bounding box, compared to some predefined threshold in pixels. 
The center distance threshold is swept to find the precision vs. threshold plots.
The precision values for a threshold of 20 pixels are shown in the brackets of the legends in the precision plots. 
Figure \ref{fig:overallprecision} shows the precision plots for all the benchmark datasets. It is seen that DaSiamRPN outperforms the other trackers in terms of precision as well. 
In Figure~\ref{fig:uavprecision}, the precision plots for different attributes of UAV123 dataset are depicted. The results show that DaSiamRPN outperforms the other tracker in most of the challenges. 

\begin{figure}[H]
\begin{subfigure}{.48\textwidth}
  \centering
  \includegraphics[width = \textwidth]{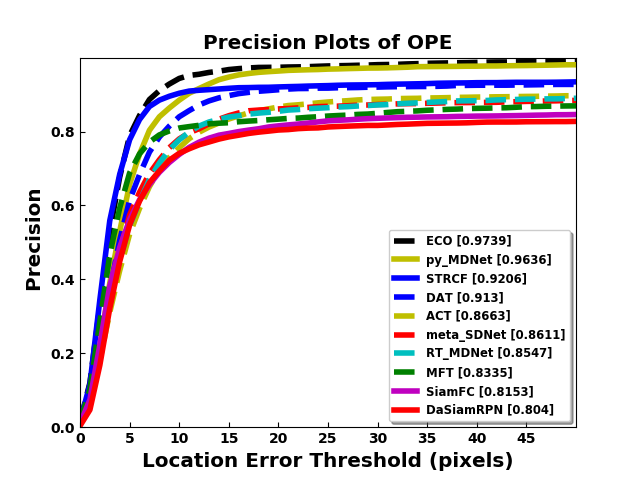}
  \caption{OTB subset}
\end{subfigure}%
\begin{subfigure}{.48\textwidth}
  \centering
  \includegraphics[width = \textwidth]{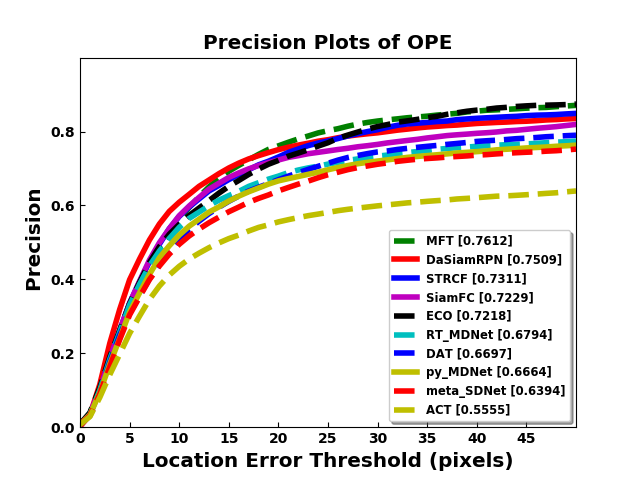}
  \caption{DTB70}
\end{subfigure} \\
\par\bigskip
\begin{subfigure}{.48\textwidth}
  \centering
  \includegraphics[width = \textwidth]{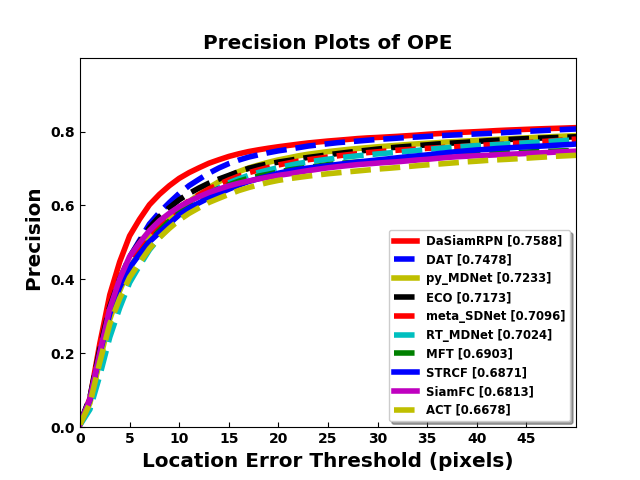}
  \caption{UAV123}
\end{subfigure}%
\begin{subfigure}{.48\textwidth}
  \centering
  \includegraphics[width = \textwidth]{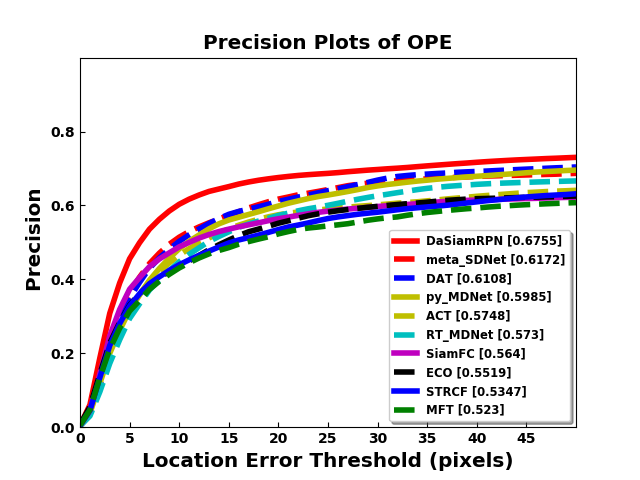}
  \caption{UAV20L}
\end{subfigure}
\caption{Precision plots of OPE.  DaSiamRPN outperforms other trackers in UAV123 and UAV 20L datasets. ECO shows best performance for the OTB subset and MFT outperform other trackers in the DTB70 datset. Best viewed zoomed in and in color.}
\label{fig:overallprecision}
\end{figure}
\unskip

\begin{figure}[H]
\centering
  \begin{tabular}{@{}cccc@{}}
    \includegraphics[width=.32\textwidth]{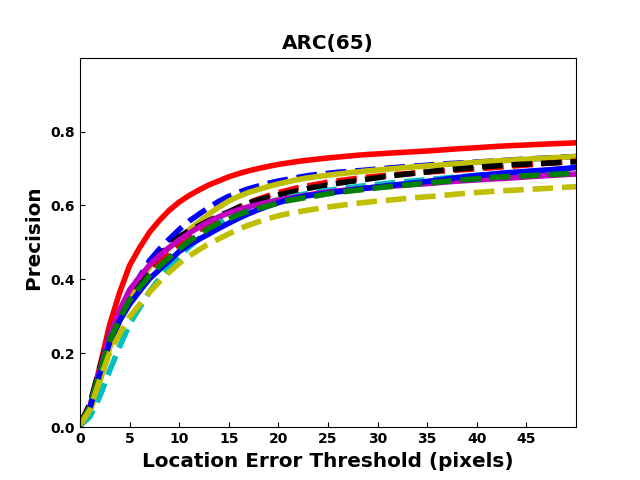} &
    \includegraphics[width=.32\textwidth]{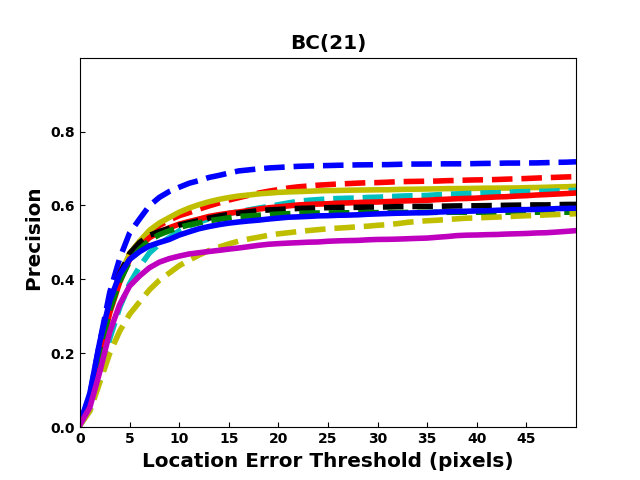} &
    \includegraphics[width=.32\textwidth]{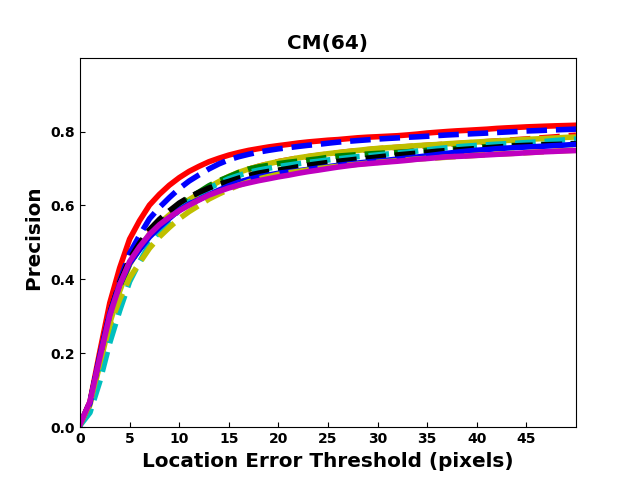} \\
    \includegraphics[width=.32\textwidth]{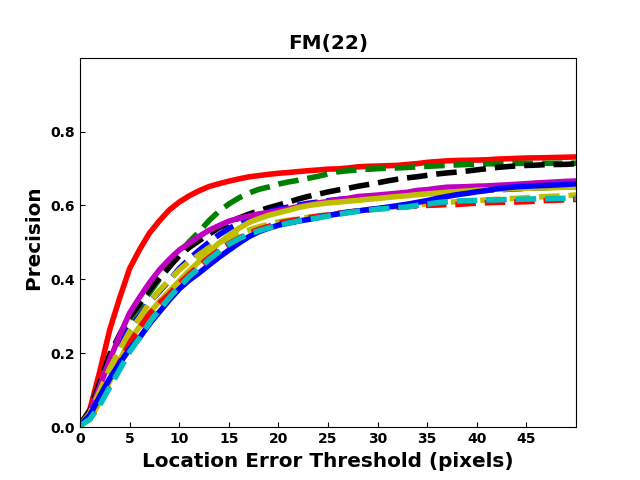} &
    \includegraphics[width=.32\textwidth]{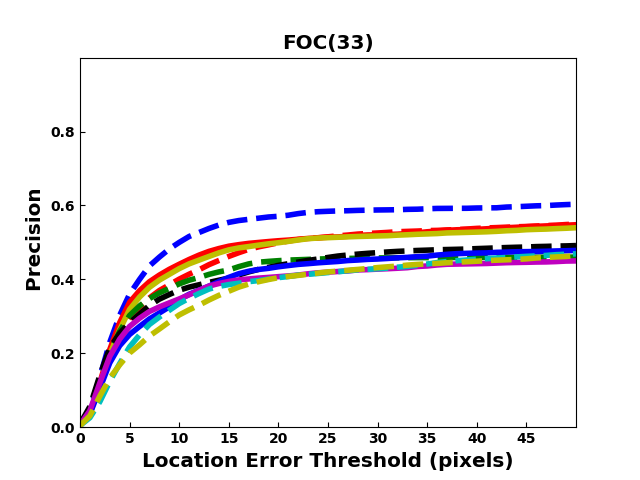} &
    \includegraphics[width=.32\textwidth]{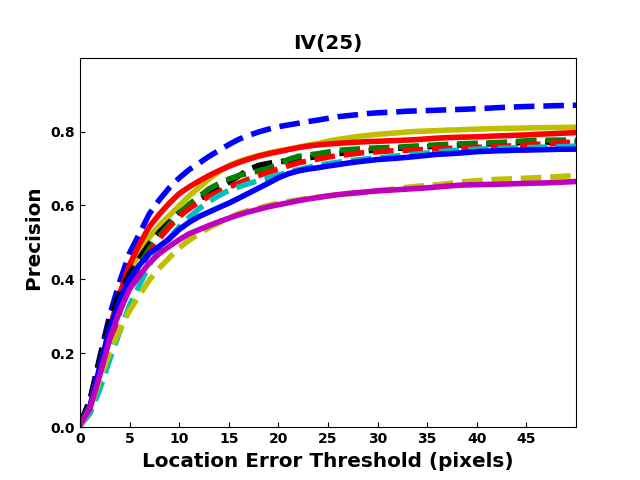} \\
      \end{tabular}
     \caption{Cont.}
\end{figure}

\begin{figure}[H]\ContinuedFloat
\centering
 \begin{tabular}{@{}cccc@{}}
    \includegraphics[width=.32\textwidth]{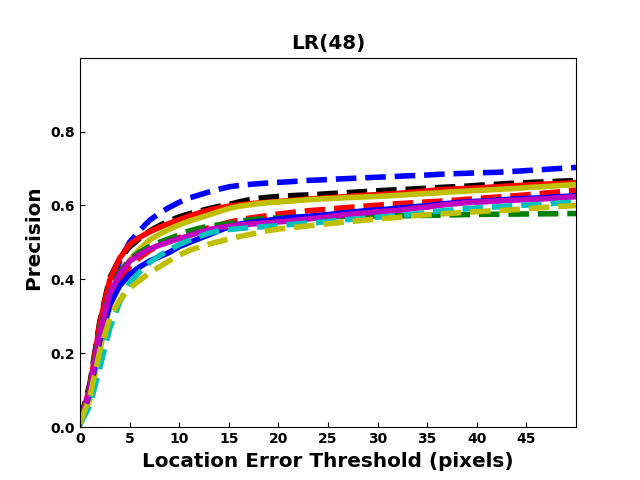} &
    \includegraphics[width=.32\textwidth]{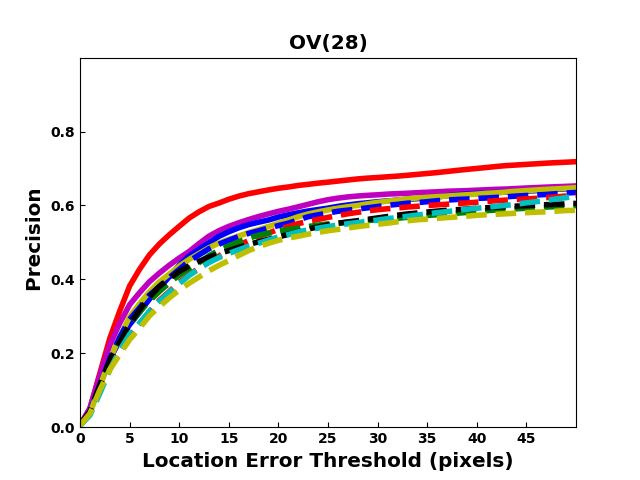}  &
    \includegraphics[width=.32\textwidth]{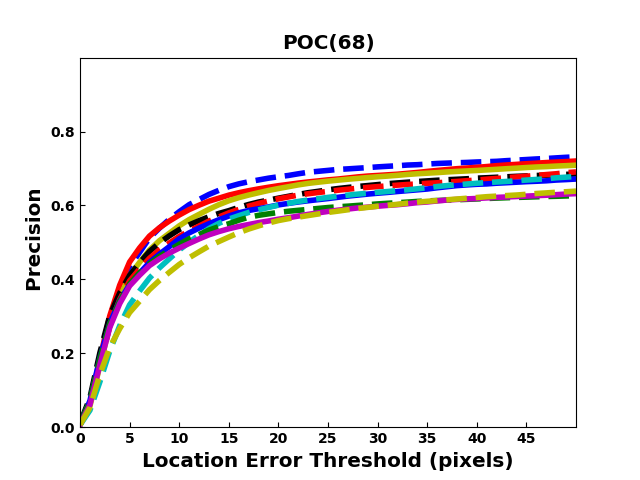} \\
    \includegraphics[width=.32\textwidth]{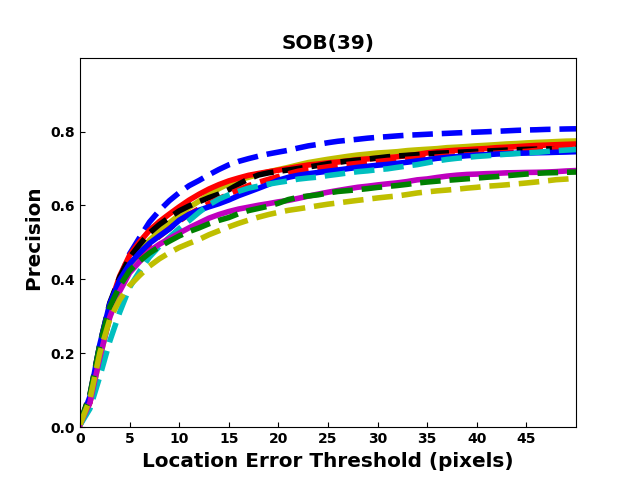} &
    \includegraphics[width=.32\textwidth]{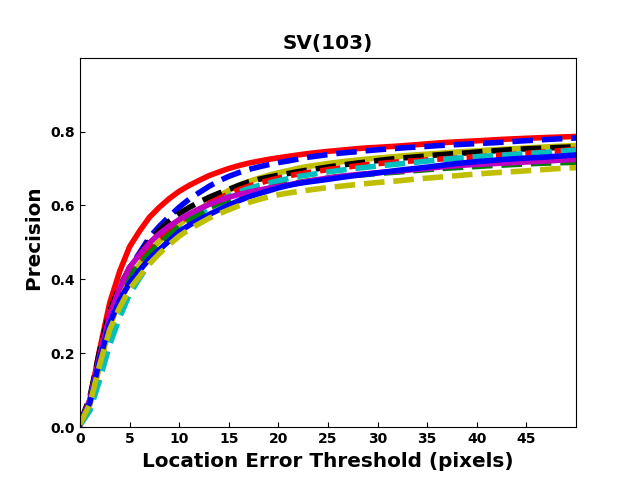} &
    \includegraphics[width=.32\textwidth]{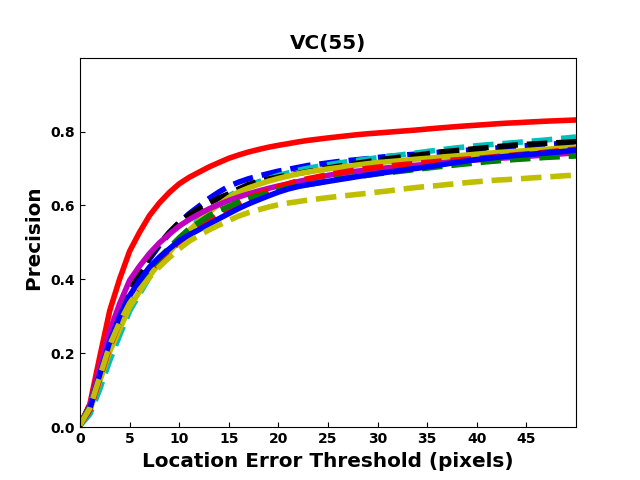}   \\
  \end{tabular}
   \includegraphics[width=.98\textwidth]{uav123low_attributes/cropped_legend.PNG}
  \caption{Precision plots of OPE of the UAV123 dataset attributes. DaSiamRPN outperforms the other trackers in most of the challenges where DAT outperforms other trackers in BC, FOC, IV, LR, and SOB challenges. Best viewed in color.}
  \label{fig:uavprecision}
\end{figure}
\unskip

\begin{table}[H]
\caption{Comparison of AUC tracker performance for various datasets. Scores show that trackers perform worse in aerial datasets compared to the ground based datasets. Red indicates top performance and blue indicates runner up. Best viewed in color.} 
\label{tab:overallcomparison}
\centering
{\begin{tabular}{ccccccc} 
\toprule
\textbf{Tracker} & \textbf{OTB 50} & \textbf{OTB 100} & \textbf{OTB Aerial} & \textbf{DTB 70} & \textbf{UAV123} & \textbf{UAV 20L} \\
\midrule
ACT & 0.657 & 0.625 & 0.6092 & 0.3784 & 0.471 & 0.4211 \\

DaSiamRPN & N/A & N/A & 0.5865 & \color{red}{0.5286} & \color{red}{0.5739} & \color{red}{0.5466}   \\

DAT & 0.704 & 0.668 & 0.6541 & 0.4413 & \color{blue}{0.5372} & 0.4566 \\

ECO & N/A & \color{red}{0.70} & \color{red}{0.7684} & \color{blue}{0.5072} & 0.5298 & \color{red}{0.5466} \\

meta-SDNet & N/A & 0.662 & 0.613 & 0.433 & 0.5236 & \color{blue}{0.4822} \\

MFT & \color{red}{0.726} & \color{blue}{0.698} & 0.6625 & 0.4957 & 0.5154 & 0.4155 \\

py-MDNet & \color{blue}{0.708} & 0.678 & 0.681 & 0.4351 & 0.5184 & 0.444 \\

RT-MDNet & N/A & 0.650 & 0.6155 & 0.4488 & 0.5012 & 0.4368 \\

SiamFC & 0.612 & N/A & 0.5961 & 0.475 & 0.4947 & 0.4255 \\

STRCF & N/A & 0.683 & \color{blue}{0.7542} & 0.5003 & 0.5132 & 0.4369 \\
\bottomrule
\end{tabular}}
\end{table} 

In Table \ref{tab:uavoverlap}, the AUC of overlap success is shown for various attributes present in the UAV123 dataset.  
In Table \ref{tab:uavprecision}, the precision at 20 pixel threshold is shown for the UAV123 dataset attributes. 
Both DaSiamRPN and DAT trackers perform well in these challenges.
In Table \ref{tab:dtboverlap}, the AUC of overlap success is provided for various attributes in the DTB70 dataset.   
In Table \ref{tab:dtbprecision}, the precision at the 20 pixel threshold is provided for different challenges present in the DTB70 dataset. DaSiamRPN achieved better results in terms of both overlap success and precision in most of the challenges of the DTB70 dataset. 

\begin{figure}[H]
    \centering
    \includegraphics[width=.98\textwidth]
    {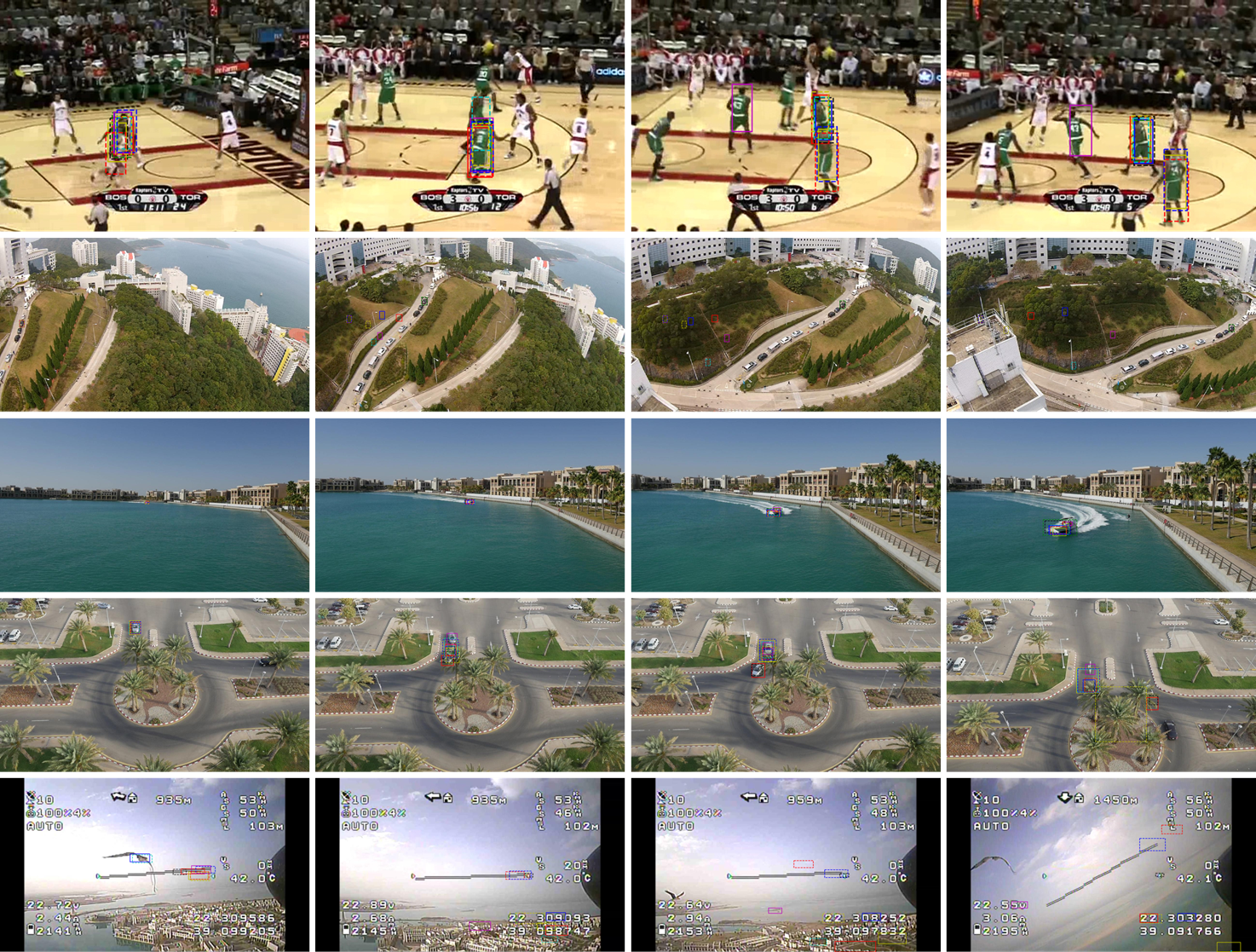}
    \includegraphics[width=.98\textwidth]{uav123low_attributes/cropped_legend.PNG}
    \caption{Tracking visualization on different sequences (top to bottom): Basketball (OTB), Car2 (DTB70), Boat8 (UAV123), Car7 (UAV123), and bird1 (UAV20L). Best viewed in color after zooming in.}
    \label{fig:visualresult}
\end{figure}

\begin{figure}[H]
    \centering 
    \includegraphics[width=0.75\textwidth]{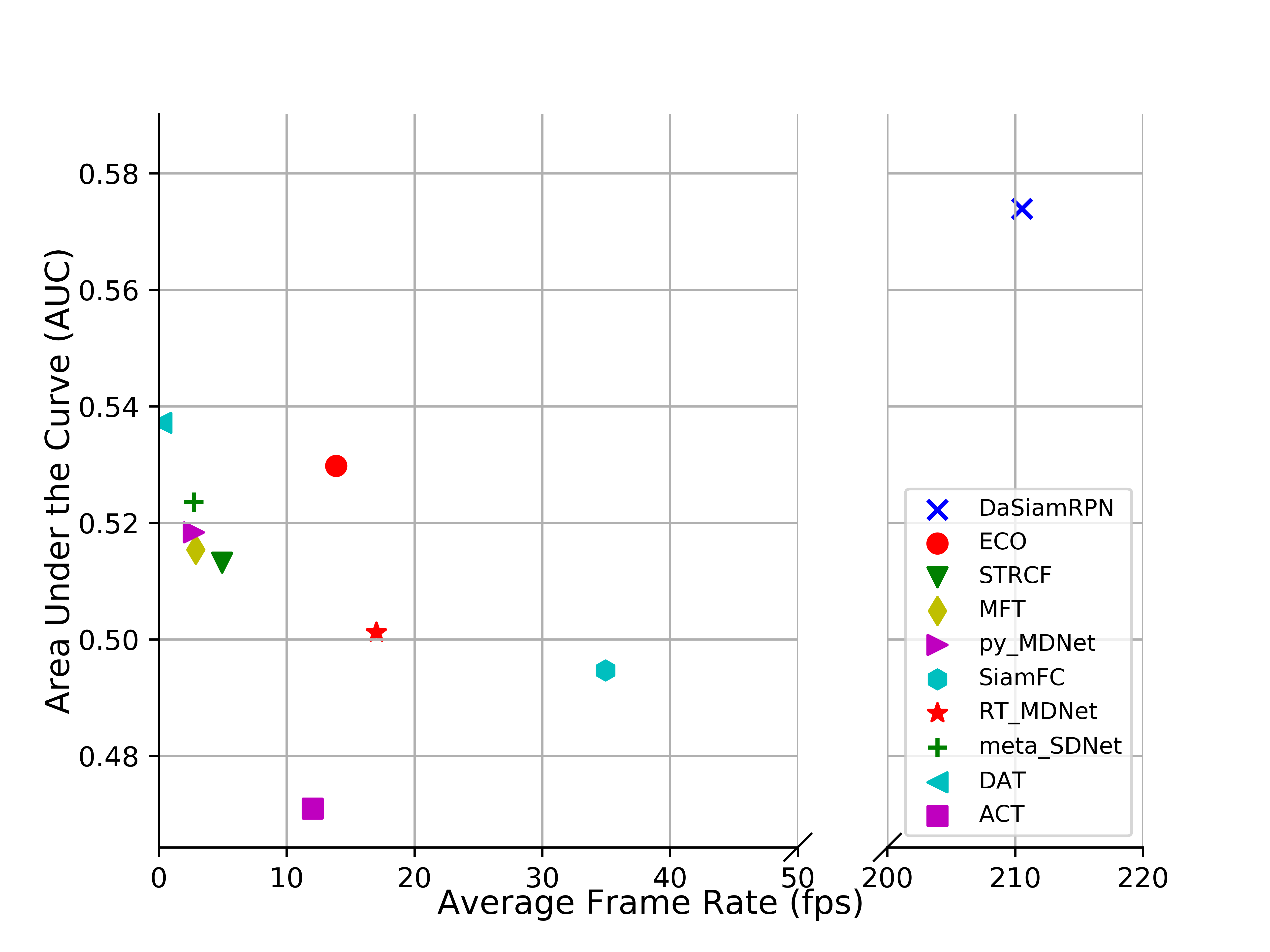}
    \caption{Results showing AUC vs. Frame Rate on UAV123 dataset. Top right corner indicates fastest implementation and best performance. Best viewed in color.
    }
    \label{fig:time}
\end{figure}

Visual results on the aerial datasets are shown in Figure \ref{fig:visualresult}. These illustrate the challenges for trackers due to the small target size, image rotation change in zoom level, and presence of similar looking distractors. 
The speed comparison is provided in Figure \ref{fig:time} on the UAV123 dataset. The fastest tracker is DaSiamRPN which runs above 200 fps and the slowest tracker is DAT which runs below 1 fps. The ECO and DAT trackers achieve good tracking performance at lower speeds. 

\begin{table}[H]
\caption{Overlap results for various UAV123 dataset attributes listed in Table \ref{tab:attributes}. Top performing tracker is shown in red and the second best performer is shown in blue. 
Best viewed in color.} 
\label{tab:uavoverlap}
\centering
\scalebox{0.76}[0.76]{
\begin{tabular}{cccccccccccccc} 
\toprule
\textbf{Tracker} & \textbf{ARC(65) }& \textbf{BC(21) }& \textbf{CM(64) }& \textbf{FM(22)} & \textbf{FOC(33)} & \textbf{IV(25)} & \textbf{LR(48)} & \textbf{OV(28)} & \textbf{POC(68)} & \textbf{SOB(39) }& \textbf{SV(103)} & \textbf{VC(55)} \\
\midrule
ACT & 0.3993 & 0.3382 & 0.4865 & 0.3677 & 0.259 & 0.4151 & 0.3126 & 0.3829 & 0.3831 & 0.4116 & 0.4404 & 0.4333\\

DaSiamRPN & \color{red}{0.527} & 0.4168 & \color{red}{0.5794} & \color{red}{0.5024} & \color{blue}{0.349} & \color{red}{0.5298} & \color{red}{0.4076} & \color{red}{0.5235} & \color{red}{0.4851} & 0.5032 & \color{red}{0.5477} & \color{red}{0.5836}  \\

DAT & \color{blue}{0.4627} & \color{red}{0.4577} & \color{blue}{0.544} & 0.3972 & \color{red}{0.3604} & \color{blue}{0.5229} & \color{blue}{0.4068} & 0.4297 & \color{blue}{0.4741} & \color{red}{0.5277} & \color{blue}{0.5109} & 0.4929 \\

ECO & 0.4594 & 0.3898 & 0.5167 & 0.4321 & 0.2873 & 0.484 & 0.3936 & 0.4101 & 0.4565 & \color{blue}{0.5098} & 0.5044 & \color{blue}{0.4953}\\

meta-SDNet & 0.4596 & \color{blue}{0.4346} & 0.5334 & 0.3667 & 0.3379 & 0.4696 & 0.3737 & 0.429 & 0.4566 & 0.5043 & 0.495 & 0.4784\\

MFT & 0.4539 & 0.4047 & 0.5259 & \color{blue}{0.4372} & 0.2983 & 0.4988 & 0.3646 & 0.4257 & 0.4329 & 0.4644 & 0.4885 & 0.4827 \\

py-MDNet & 0.4569 & 0.4139 & 0.5206 & 0.3805 & 0.3206 & 0.4855 & 0.376 & 0.4271 & 0.4548 & 0.4928 & 0.4906 & 0.4723\\

RT-MDNet & 0.4198 & 0.3876 & 0.5042 & 0.3607 & 0.2655 & 0.4558 & 0.3407 & 0.4058 & 0.4243 & 0.4739 & 0.4716 & 0.4685 \\

SiamFC & 0.427 & 0.3336 & 0.4977 & 0.4113 & 0.2713 & 0.4143 & 0.3566 & 0.4457 & 0.4048 & 0.439 & 0.4694 & 0.4627\\

STRCF & 0.4508 & 0.3761 & 0.5231 & 0.3863 & 0.2774 & 0.4632 & 0.3456 & \color{blue}{0.446} & 0.4408 & 0.4895 & 0.4819 & 0.4828 \\
\bottomrule
\end{tabular}}
\end{table} 
\unskip

\begin{table}[H]
\caption{Precision results for various UAV123 dataset attributes listed in Table \ref{tab:attributes}. Top performing tracker is shown in red and the second best performer is shown in blue. 
Best viewed in color.} 
\label{tab:uavprecision}
\centering
\scalebox{0.76}[0.76]{\begin{tabular}{cccccccccccccc} 
\toprule
\textbf{Tracker} & \textbf{ARC(65) }& \textbf{BC(21) }& \textbf{CM(64) }& \textbf{FM(22)} & \textbf{FOC(33)} & \textbf{IV(25)} & \textbf{LR(48)} & \textbf{OV(28)} & \textbf{POC(68)} & \textbf{SOB(39) }& \textbf{SV(103)} & \textbf{VC(55)} \\
\midrule
ACT & 0.572 & 0.5233 & 0.6841 & 0.5557 & 0.4046 & 0.6058 & 0.5361 & 0.5056 & 0.5591 & 0.5814 & 0.6289 & 0.602 \\

DaSiamRPN & \color{red}{0.7108} & 0.5962 & \color{red}{0.7619} & \color{red}{0.687} & \color{blue}{0.5045} & 0.7454 & 0.6108 & \color{red}{0.6465} & \color{blue}{0.6534} & 0.6955 & \color{red}{0.7289} & \color{red}{0.7627}  \\

DAT & \color{blue}{0.6661} & \color{red}{0.7028} & \color{blue}{0.7531} & 0.5904 & \color{red}{0.5706} & \color{red}{0.8129} & \color{red}{0.6625} & 0.5455 & \color{red}{0.6776} & \color{red}{0.7444} & \color{blue}{0.716} & \color{blue}{0.6928} \\

ECO & 0.6287 & 0.5891 & 0.6957 & 0.6012 & 0.4379 & 0.7179 & \color{blue}{0.6242} & 0.5151 & 0.6197 & 0.6921 & 0.6817 & 0.6768\\

meta-SDNet & 0.6347 & \color{blue}{0.6427} & 0.7187 & 0.5522 & 0.499 & 0.6984 & 0.5767 & 0.5352 & 0.618 & 0.6783 & 0.6741 & 0.6507 \\

MFT & 0.6069 & 0.5767 & 0.713 & \color{blue}{0.6577} & 0.4506 & 0.7111 & 0.5664 & 0.5308 & 0.5811 & 0.6061 & 0.6521 & 0.6403 \\

py-MDNet & 0.6583 & 0.6353 & 0.7186 & 0.58 & 0.499 & \color{blue}{0.7476} & 0.6096 & 0.5516 & 0.6457 & \color{blue}{0.697} & 0.6884 & 0.6731\\

RT-MDNet & 0.6158 & 0.6019 & 0.7052 & 0.5459 & 0.4047 & 0.6826 & 0.5451 & 0.5149 & 0.6007 & 0.6623 & 0.6669 & 0.6812 \\

SiamFC & 0.6143 & 0.4967 & 0.6774 & 0.5861 & 0.4073 & 0.602 & 0.5555 & \color{blue}{0.5841} & 0.5627 & 0.6098 & 0.6515 & 0.6527\\

STRCF & 0.6075 & 0.5657 & 0.6871 & 0.5466 & 0.4341 & 0.6747 & 0.5636 & 0.5686 & 0.6014 & 0.6683 & 0.6476 & 0.6361 \\
\bottomrule
\end{tabular}}
\end{table} 
\unskip

\begin{table}[H]
\caption{Overlap results for various DTB70 dataset attributes listed in Table \ref{tab:attributes}. Top performing tracker is shown in red and the second best performer is shown in blue. 
Best viewed in color.} 
\label{tab:dtboverlap}
\centering
\scalebox{0.8}[0.8]{\begin{tabular}{ccccccccccccc} 
\toprule
\textbf{Tracker} & \textbf{SV(22)} & \textbf{ARV(25)} & \textbf{OCC(17)} & \textbf{DEF(18)} & \textbf{FCM(41)} & \textbf{IPR(47)} & \textbf{OPR(6)} & \textbf{OV(7)} & \textbf{BC(13)} & \textbf{SOA(27)} & \textbf{MB(27)} \\
\midrule
ACT & 0.3439 & 0.3507 & 0.3995 & 0.3210 & 0.3534 & 0.3261 & 0.2351 & 0.3661 & 0.2715 & 0.3420 & 0.3089 \\

DaSiamRPN & \color{red}{0.5905} & \color{red}{0.5339} & 0.4203 & \color{red}{0.5502} & \color{red}{0.5240} & \color{red}{0.5013} & \color{red}{0.4337} & \color{red}{0.5008} & 0.4402 & 0.4514 & 0.4668   \\

DAT & 0.4483 & 0.4280 & 0.3910 & 0.4489 & 0.4361 & 0.4384 & 0.3589 & \color{blue}{0.4665} & 0.3308 & 0.3835 & 0.3652 \\

ECO & 0.4753 & 0.4344 & \color{red}{0.5121} & 0.4477 & \color{blue}{0.5202} & 0.4584 & 0.3257 & 0.4318 & 0.4434 & \color{red}{0.5255} & \color{red}{0.5090} \\

meta-SDNet & 0.4508 & 0.3993 & 0.4045 & 0.4630 & 0.4090 & 0.3996 & 0.3162 & 0.3949 & 0.2989 & 0.4029 & 0.3322 \\

MFT & \color{blue}{0.5362} & \color{blue}{0.4475} & 0.4324 & \color{blue}{0.5004} & 0.5080 & \color{blue}{0.4897} & \color{blue}{0.4239} & 0.4019 & \color{blue}{0.4614} & \color{blue}{0.4849} & \color{blue}{0.5073} \\

py-MDNet & 0.4286 & 0.3900 & 0.4305 & 0.4334 & 0.4314 & 0.4176 & 0.3500 & 0.4643 & 0.3527 & 0.3815 & 0.3655 \\

RT-MDNet & 0.4254 & 0.3777 & \color{blue}{0.4878} & 0.3243 & 0.4948 & 0.4011 & 0.3175 & 0.3999 & 0.3083 & 0.4694 & 0.4035 \\

SiamFC & 0.4764 & 0.4280 & 0.4257 & 0.4311 & 0.4974 & 0.4615 & 0.3701 & 0.4159 & \color{red}{0.4663} & 0.4600 & 0.4736 \\

STRCF & 0.5156 & 0.4388 & 0.4714 & 0.4979 & 0.5181 & 0.4718 & 0.3550 & 0.4356 & 0.4107 & 0.4827 & 0.4797 \\
\bottomrule
\end{tabular}}
\end{table} 
\unskip

\begin{table}[H]
\caption{Precision results on various DTB70 dataset attributes listed in Table \ref{tab:attributes}. Top performing tracker is shown in red and the second-best performing tracker is shown in blue. Best viewed in color.} 
\label{tab:dtbprecision}
\centering    
\scalebox{0.8}[0.8]{\begin{tabular}{ccccccccccccc} 
\toprule
\textbf{Tracker} & \textbf{SV(22)} & \textbf{ARV(25)} & \textbf{OCC(17)} & \textbf{DEF(18)} & \textbf{FCM(41)} & \textbf{IPR(47)} & \textbf{OPR(6)} & \textbf{OV(7)} & \textbf{BC(13)} & \textbf{SOA(27)} & \textbf{MB(27)} \\
\midrule
ACT & 0.4434 & 0.4658 & 0.6437 & 0.4245 & 0.5484 & 0.4707 & 0.2138 & 0.5688 & 0.4229 & 0.5321 & 0.4892 \\

DaSiamRPN & \color{red}{0.7749} & \color{red}{0.7363} & 0.6233 & \color{red}{0.7774} & 0.7465 & \color{blue}{0.7086} & \color{blue}{0.5252} & 0.6950 & 0.6527 & 0.6638 & 0.6811  \\

DAT & 0.6783 & 0.6451 & 0.5823 & 0.6978 & 0.6586 & 0.6501 & 0.4935 & \color{red}{0.7192} & 0.5316 & 0.6043 & 0.5694 \\

ECO & 0.5982 & 0.6007 & \color{blue}{0.7449} & 0.6052 & 0.7523 & 0.6376 & 0.3210 & 0.5851 & 0.7196 & \color{red}{0.8023} & \color{blue}{0.7362} \\

meta-SDNet & 0.6224 & 0.5751 & 0.6272 & 0.6545 & 0.5963 & 0.5771 & 0.3883 & 0.5802 & 0.4336 & 0.6371 & 0.4851 \\

MFT & \color{blue}{0.7517} & \color{blue}{0.6463} & 0.7074 & 0.7155 & \color{red}{0.7922} & \color{red}{0.7551} & 0.5144 & 0.6315 & \color{red}{0.8332} & \color{blue}{0.7807} & \color{red}{0.8237} \\

py-MDNet & 0.6299 & 0.5577 & 0.6778 & 0.6628 & 0.6624 & 0.6284 & 0.4605 & \color{blue}{0.7110} & 0.5665 & 0.6093 & 0.5816 \\

RT-MDNet & 0.6017 & 0.5512 & \color{red}{0.7539} & 0.4828 & 0.7558 & 0.5924 & 0.4378 & 0.6247 & 0.5065 & 0.7135 & 0.6315 \\

SiamFC & 0.6800 & 0.6407 & 0.6780 & 0.6392 & 0.7626 & 0.7004 & \color{red}{0.5309} & 0.6713 & \color{blue}{0.7552} & 0.7184 & 0.7348 \\

STRCF & 0.7079 & 0.6371 & 0.7048 & \color{blue}{0.7440} & \color{blue}{0.7700} & 0.6904 & 0.4970 & 0.6572 & 0.6587 & 0.7318 & 0.7230 \\
\bottomrule
\end{tabular}}
\end{table} 

\unskip

\subsection{Overall Comparison}
The overall quantitative comparison of the trackers is shown in Table \ref{tab:overallcomparison} for all benchmark datasets. 
The results for the OTB aerial subset are slightly worse than the overall OTB results. 
As the OTB subset does not exhibit all the attributes of the other aerial datasets, tracker performance remains similar with respect to the overall ground level OTB dataset. 
The evaluation plots in Figures \ref{fig:overallsuccess} and \ref{fig:overallprecision} showed that the ECO tracker performs the best and STRCF is the second best for the OTB subset. 

Tracker performance degrades consistently for the aerial datasets compared to ground level tracking.
For the DTB70 dataset, the DaSiamRPN and MFT trackers perform the best for overlap and precision respectively. 
The DaSiamRPN tracker performs the best for the UAV123 dataset and UAV20L datasets. The distractor aware training, region proposal network for better IoU, and local to global search for redetection make the DaSiamRPN the best overall performer for aerial tracking. 

\subsection{Attribute Comparison}
Attributes are annotated in the datasets, as shown in Table \ref{tab:attributes}, and are used to evaluate the tracker performance under various challenging conditions. 
In different sequences, one or several attributes may be present.
The evaluation of trackers based on specific video attributes is shown in \mbox{Figures \ref{fig:uavsuccess} and \ref{fig:uavprecision}} and Tables \ref{tab:uavoverlap} and \ref{tab:uavprecision} for  UAV123 and in Tables \ref{tab:dtboverlap} and  \ref{tab:dtbprecision} for DTB70. 
These results provide insights in the performance of the trackers under various conditions.

The aspect ratio change (ARC-UAV123) and the aspect ratio variation (ARV-DTB70) attributes specify the target's aspect ratio change over the temporal span of the sequence. 
The evaluation shows that the DaSiamRPN outperforms all the other trackers on this challenge. The region proposal network helps DaSiamRPN achieve better performance in terms of aspect ratio change. Note that in this specific challenge, in the UAV123 dataset, CF-based trackers and MDNet-based trackers perform similarly, whereas CF-based trackers perform relatively well compared to the MDNet-based trackers in the DTB70 dataset. We conclude that the model update strategy of the CF-based trackers help to achieve relatively better performance compared to he MDNet-based trackers.     

The Background Clutter (BC) attribute is present in both datasets. This attribute specifies instances where the target is hardly distinguishable from the background. Based on the evaluation, DAT outperforms the other trackers for UAV123 and SiamFC does best for DTB70, for both overlap and precision. Note that SiamFC performs the worst for UAV123 sequences where BC is present. 

Camera motion (CM-UAV123), fast motion (FM-UAV123), and fast camera motion (FCM-DTB70) attributes are evaluated. These attributes specify faster relative speed between the camera and the target. The DaSiamRPN tracker outperforms the other trackers for both datasets. However, MDNet based trackers performed comparatively well in the UAV123 dataset when camera motion is involved. However, when the relative motion is fast, the CF-based trackers outperform the MDNet based trackers in both datasets. We attribute this to the better model update strategy of the correlation filter based trackers. 

Full occlusion (FOC-UAV123), partial occlusion (POC-UAV123), and Occlusion (OCC-DTB70) attributes are also evaluated. In POC, the target becomes partially occluded and then reappears without full occlusion, whereas in FOC, the target may be fully occluded for several frames and then reappears. The OCC attribute in DTB70  addresses both full and partial occlusion. It is important for the tracker to have redetection capabilities and stop updating the appearance model during full or partial occlusion to prevent drift from the target. Based on the tracker evaluation on these datasets, we~find that for full occlusion DAT performed well, whereas for partial occlusion, DaSiamRPN performed well. For OCC in the DTB70 dataset, ECO performed best. 

The Out of View (OV) attribute is present in both datasets. This attribute specifies that the target is no longer within the field of view of the camera, which is particularly challenging for most trackers. To~do well in this challenge, the tracker must have redetection capabilities and be discriminative enough to distinguish the target from other similar looking objects. DaSiamRPN outperforms all the other trackers in this challenge, because it incorporates local-to-global search strategy.  

The attributes Similar Object (SOB-UAV123) and Similar Object Around (SOA-DTB70) indicate that objects with shape and appearance similar to the target appear near the target. These objects are also called distractors. Tracking is more challenging when the target is partially occluded and distractors are present close to the target. The object may also be fully occluded by the distractor. For~this challenge, the results show that CF-based trackers perform well in the DTB70 dataset, whereas MDNet-based trackers perform well in the UAV123 dataset. 
The performance of the Siamese trackers is comparatively lower because the correlation map generally provides high scores on the distractors, which sometimes causes tracker failure. 

The Scale Variation (SV) attribute is present in both datasets. It specifies those sequences where the scale of the target changes over time. The results show that DaSiamRPN is significantly better than other trackers. Again, we attribute this to the Region Proposal Network architecture of the DaSiamRPN tracker. Illumination variation (IV), low resolution (LR), and viewpoint change (VC) are some other attributes that are present in the UAV123 dataset. 
IV specifies the sequences where illumination change is involved related to the target and LR specifies those sequences where the target has low resolution. 
VC specifies the changes in the camera viewpoint over the temporal span of a sequence. In terms of overlap, DaSiamRPN outperforms the other trackers in these challenges. However, for the precision, MDNet outperforms the other trackers for LR and IV attributes. 

Deformation (DEF), In-plane Rotation (IPR), Out of Plane Rotation (OPR), and Motion Blur (MB) are some other attributes present in the DTB70 dataset. Deformation specifies the shape change of the object, in-plane and out-of-plane rotation specifies whether the target object is rotating inside or outside from the image plane, and motion blur specifies blurred target during tracking. For deformation, out-of-plane rotation, and in-plane-rotation, DaSiamRPN performs best. For MB, ECO performs best among the compared trackers for overlap success. However, for the precision success, MFT performs best. The results show that trackers perform much better in the IPR challenges compared to the OPR challenges. In presence of the MB attribute, CF-based trackers generally performs better compared to the other trackers. 

\subsection{Visual Comparison}
We  present visual examples of the results for all trackers  in Figure \ref{fig:visualresult}. 
The Basketball sequence is taken from the OTB dataset.  Although it is not an aerial image, it contains distractors that are large enough to observe.
Partial occlusion and distractor attributes are present in this sequence. 
SiamFC and some MDNet based trackers lost the target by the end of the sequence. 

The Car2 sequence from DTB70 dataset specifies the camera motion where the camera rotates around the target. The results show that eight out of 10 trackers lost track as soon as there is significant rotation of the camera in this sequence. 

The Boat8 and Car7 sequences are taken from the UAV123 dataset. Boat8 has significant scale and aspect ratio change where the appearance of the target also changed. These results show that ECO and MFT trackers perform well but eventually drift. Surprisingly, DaSiamRPN got stuck on small part of the object, most likely due to a proposal generated for that region. 
In Car7, the target was occluded by a tree while another distractor appeared at the same time. Only the ECO and the DaSiamRPN trackers were able to successfully handle the situation, whereas all other trackers started to track the distractor. In cases of full occlusion, all the trackers lost the target. 

Finally, the bird1 sequence from UAV123 dataset is evaluated. Here the target is moving fast and went out-of-view multiple times for several frames. It is seen that due to the small target size, fast motion and partial occlusion, only STRCF and RT-MDNet successfully track the object until it goes out-of-view, but no tracker can redetect the target when it reappears.

\subsection{Speed Comparison}
The speed comparison among all the trackers is shown in Figure \ref{fig:time} where the AUC vs. frame rate is plotted for the UAV123 dataset. SN-based trackers have higher frame rate compared to the other trackers. This is because the network parameters are not updated during online tracking. The~DaSiamRPN achieves significantly higher frame rate because of its approach to online tracking as one shot learning. Among the CF-based trackers, ECO has the highest frame rate and outperforms the other CF-based trackers. The factorized convolution operation makes the tracker more efficient, thereby allowing it to achieve a higher frame rate and better performance. Among the MDNet based trackers, RT-MDNet achieves real-time performance with a small accuracy drop from the the original MDNet tracker. 
It is also seen that DAT performs well, but it has the lowest frame rate due to the update strategy of the tracker where the tracker updates on all the frames with score lower than a certain threshold.

\section{Conclusions}
\label{sec:conclusion}
In this study, we benchmarked ten state-of-the-art CNN-based visual object trackers from four different classes: Siamese Network-based, Tracking by Detection-based, Correlation Filter-based, and Reinforcement Learning.
We considered four datasets: a subset of OTB, DTB70, UAV123, and UAV20L datasets for testing and comparing the tracking algorithms. 
Visual examples of different trackers are shown and the results of an One Pass Evaluation (OPE) are reported. We compared the results among different datasets as well as specific attribute challenges within the datasets. 
Trackers performed worse in the aerial datasets than in the typical ground level videos. 

In our study, we found that  Siamese network based trackers face difficulty when there are distractors present within the sequence. This is because the cross-correlation operation will create strong peaks on the distractors and drift may occur particularly when the main target is occluded. Siamese trackers do not have any online model update, which makes them fast but occasionally cannot handle significant appearance change of the object. 
However, DaSiamRPN performs well due to distractor aware training and accurate localization based on the RPN.
The challenge for the CF-based trackers is to find a proper update strategy such that the tracker does not update the appearance model when the target is absent. 
Among MDNet based trackers, py-MDNet and DAT are computationally expensive where RT-MDNet and meta-tracker run at higher frame rate with relatively lower accuracy. 
Finally, the RL-based tracker is yet to achive the desired accuracy.

It is notable that none of the trackers is designed for reidentification, which is needed for target reacquisition when the target goes out of view and reappears. 

The overall performance of the implemented trackers indicates that further research is needed to reach the full potential of deep learning tracking on aerial sequences.


\vspace{6pt} 



\authorcontributions{
A.M.N.T. contributed in performing the experiments and writing the report.
B.M. contributed in conceptualization.
A.S. contributed in conceptualization, review and editing, and supervising the project. All authors have read and agreed to the published version of the manuscript.
}

\funding{
This research was supported in part by the Air Force Research Laboratory, Sensors Directorate (AFRL/RYAP) under contract number FA8650-18-C-1739 to Systems and Technology Research. 
}



\conflictsofinterest{The authors declare no conflicts of interest. The funders helped reviewing the manuscript and provided the decision to publish the result. 

} 

\reftitle{References}





\end{document}